%% file: ms.tex
\documentclass[acmsmall,nonacm,screen]{acmart}
\pdfoutput=1

\usepackage[utf8]{inputenc}
\usepackage{subcaption}
\usepackage{hyperref}
\usepackage{tabularx}
\usepackage{algorithm2e}
\usepackage{multirow}

\setcopyright{none}

\newcommand\independent{\protect\mathpalette{\protect\independenT}{\perp}}
\def\independenT#1#2{\mathrel{\rlap{$#1#2$}\mkern2mu{#1#2}}}

\title{A Survey of Methods, Challenges and Perspectives in Causality}

\author{Ga\"{e}l Gendron}
\orcid{0000-0002-2457-934X}
\affiliation{
    \institution{University of Auckland}
    \city{Auckland}
    \country{New Zealand}
    }
\email{ggen187@aucklanduni.ac.nz}

\author{Michael Witbrock}
\orcid{0000-0002-7554-0971}
\affiliation{
    \institution{University of Auckland}
    \city{Auckland}
    \country{New Zealand}
    }
\email{m.witbrock@auckland.ac.nz}

\author{Gillian Dobbie}
\orcid{0000-0001-7245-0367}
\affiliation{
    \institution{University of Auckland}
    \city{Auckland}
    \country{New Zealand}
    }
\email{g.dobbie@auckland.ac.nz}

\begin{document}

\input{abstract}

\maketitle
\renewcommand{\shortauthors}{Gendron, Witbrock and Dobbie}

\input{introduction}

\input{causality_theory}

\input{neural_causal_model}

\input{causal_discovery}

\input{causal_inference}

\input{representation_learning}

\input{applications_short}

\input{conclusion}

\bibliographystyle{ACM-Reference-Format}
\bibliography{references/intro,references/graph_theory,references/causality,references/neural_causal_model,references/representation_learning,references/applications}

\newpage
\appendix

\input{applications}

\end{document}

%% file: abstract.tex
\begin{abstract}

Deep Learning models have shown success in a large variety of tasks by extracting correlation patterns from high-dimensional data but still struggle when generalizing out of their initial distribution. As causal engines aim to learn mechanisms independent from a data distribution, combining Deep Learning with Causality can have a great impact on the two fields. In this paper, we further motivate this assumption. We perform an extensive overview of the theories and methods for Causality from different perspectives, with an emphasis on Deep Learning and the challenges met by the two domains. We show early attempts to bring the fields together and the possible perspectives for the future. We finish by providing a large variety of applications for techniques from Causality.
    
\end{abstract}

%% file: introduction.tex
\section{Introduction}

One motivation for studying Causality comes from the sentence "correlation does not imply causation", meaning that it is not sufficient to have two statistically correlated variables X and Y to deduce that one is causing the other. The \textit{Common Cause Principle} \cite{reichenbach1956direction} brings restriction to this statement by saying that either X causes Y, Y causes X, or a third variable Z is the cause of both. However, this is not enough to fully characterize a causal relationship, and different theories have been developed to this end. Many fields of science aim to find causal relationships, e.g. social sciences \cite{duncan2014introduction, dundar2022causal}, economics \cite{haavelmo1943statistical}, neuroscience \cite{FRENCH20208}. This explains the great interest in the topic of Causality and the variety of approaches tackling it.

Deep Learning (DL) has been very successful in a large variety of tasks by extracting correlation patterns from high-dimensional data. However, current methods still struggle at generalizing in a distribution different from their training data. This can be explained by the presence of confounding effects in the data, leading to spurious correlations. The field of Causality studies the effects of such confounders and aims to find systematic ways to learn general causal effects invariant of the data distribution. Therefore, it can be hypothesised that adding Causality theory to DL can improve performance and generalization capabilities. However, the fields of Deep Learning and Causality developed separately, until recently when a growing interest in merging the two fields has arisen \cite{DBLP:journals/corr/abs-2204-00607}.

These observations motivate this survey of Causality theories and techniques. We look at it from the perspective of Machine Learning but attempt to cover a large spectrum of methods that are applied to various domains that do not typically intersect. We cover views of Causality for static data and time series and present the domains in which they are applied. We also describe the challenges encountered by Causality and Deep Learning and how techniques combining the two fields have been applied to solve them. Many open questions remain to be solved, we introduce them and describe the current perspectives in the research literature on these topics to answer them.
We summarize our contributions here:
\begin{itemize}
    \item We conduct an extensive survey of the field of Causality within the Structural Causal Model, broadened with the Potential Outcome Framework and Granger Causality, domains usually considered separately.
    \item We motivate the use of Causality for Deep Learning and the possible future perspectives in the field.
    \item We find and describe several connections between Causality and Deep Learning including Graph Neural Networks, sparse architectures, data augmentation, and disentanglement.
    \item We provide a detailed list of works and domains of application for Causality.
\end{itemize}

To the best of our knowledge, these contributions are the first to gather the various approaches of Causality and bring together the applications using them. We also make original connections between Causality and with Deep Learning.
In contrast, the survey by \citet{DBLP:journals/ijar/ZangaOS22} focuses on causal structure discovery and \citet{DBLP:journals/tkdd/YaoCLLGZ21} discuss only causal inference approaches.
Surveys with a broader range \cite{10.3389/fgene.2019.00524, DBLP:journals/csur/GuoCLH020, DBLP:journals/widm/NogueiraPRPG22, DBLP:journals/csur/VowelsCB23} include both tasks but do not cover the most recent Deep Learning techniques. 
Additionally, \citet{DBLP:journals/kais/MoraffahSKBWTRL21} discuss methods for time-series and \citet{DBLP:journals/corr/abs-2206-15475} provide another extensive survey of Machine Learning methods for Causality.

The rest of the paper is structured as follows: In Section \ref{sec:causality_theory}, we introduce the concepts and terminologies of the main causality theories. 
In Section \ref{sec:neural_causal_model}, we motivate and introduce Neural Causal Models, a novel framework for causal reasoning based on structural causal models and taking advantage of Deep Learning techniques. 
In Sections \ref{sec:causal_structure_discovery} and \ref{sec:causal_inference}, we perform an extensive survey of causality frameworks and methods for causal structure discovery and causal inference from a computer science perspective.
Section \ref{sec:representation_learning} introduces another causal task that has arisen with the advent of Deep Learning, called causal representation learning, and introduces the recent techniques developed in the domain. 
We present applications of Causality in various fields of science  in Section \ref{sec:applications}. We provide a conclusion to our work in Section \ref{sec:conclusion}.

\begin{table}
    \footnotesize
    \centering
    \begin{tabularx}{\linewidth}{>{\hsize=.15\linewidth}X>{\hsize=.15\linewidth}X>{\hsize=.15\linewidth}X>{\hsize=.23\linewidth}X>{\hsize=.32\linewidth}X}
        \hline
        \multicolumn{4}{>{\centering\hsize=.68\linewidth}X}{Categories and subcategories} & \multicolumn{1}{>{\centering\hsize=.32\linewidth}X}{Papers} \\
        \hline
        \multirow{6}{.18\linewidth}{Causal Structure Discovery}     & Stationary    & Graphical         & Constraints (\ref{sec:constraints}) & \cite{DBLP:conf/uai/VermaP90, spirtes1989causality, DBLP:books/daglib/0023012, DBLP:journals/jmlr/ColomboM14, DBLP:conf/uai/RamseyZS06, buhlmann2010variable, DBLP:journals/ml/TsamardinosBA06, DBLP:books/daglib/0023012, colombo-learning-2012, DBLP:conf/nips/RohekarNGN21, DBLP:conf/pgm/OgarrioSR16} \\
                                                                    &               &                   & Scores (\ref{sec:scores}) & 
        \cite{DBLP:journals/ml/CooperH92, meek1997graphical, DBLP:journals/jmlr/Chickering02a, DBLP:conf/uai/HauserB11, ramsey2017million, DBLP:conf/pgm/OgarrioSR16, DBLP:conf/nips/ZhengARX18} \\
                                                                    &               &                   & Non-Gaussian (\ref{sec:non_gaussian}) & 
        \cite{DBLP:journals/jmlr/ShimizuHHK06, zhang2010distinguishing, DBLP:conf/uai/ZhangH09, DBLP:journals/corr/abs-2204-00762} \\
                                                                    &               &                   & Variational (\ref{sec:variational_discovery}) & 
        \cite{DBLP:journals/corr/abs-2106-07635} \\
                                                                    &               &                   & LLMs (\ref{sec:llm_discovery}) &  
        \cite{DBLP:journals/corr/abs-2308-13067, DBLP:journals/corr/abs-2306-05836} \\
        \cline{2-5}
                                                                    & Time-Series   & Graphical         & Constraints (\ref{sec:constraints}) &    
        \cite{entner20120, doi:10.1126/sciadv.aau4996} \\
        \hline
        \multirow{9}{.18\linewidth}{Causal Inference}              & Stationary     & Do-Calculus       & (\ref{sec:do_calculus}) &                          
        \cite{pearl-2009, DBLP:conf/uai/HuangV06} \\
                                                                    &               & Learning          & Re-weighting (\ref{sec:re_weighting}) & 
        \cite{10.1093/biomet/70.1.41, doi:10.1080/01621459.1994.10476818, https://doi.org/10.1111/rssb.12027} \\
                                                                    &               &                   & Matching (\ref{sec:matching}) & 
        \cite{doi:10.1080/00273171.2011.568786, 10.1093/biomet/70.1.41, stuart2010matching, king-nielsen-2019, 10.5555/3061053.3061146, iacus-king-porro-2012, DBLP:journals/tkdd/YaoCLLGZ21} \\
                                                                    &               &                   & Decision Trees (\ref{sec:trees}) & 
        \cite{https://doi.org/10.1002/widm.8, 10.1214/09-AOAS285} \\
                                                                    &               &                   & NCMs (\ref{sec:ncm_inference}) &  
        \cite{DBLP:journals/corr/abs-2110-12052, DBLP:journals/corr/abs-2109-04173} \\
                                                                    &               &                   & Variational (\ref{sec:vgae}) & 
        \cite{DBLP:journals/corr/KipfW16a, DBLP:journals/corr/abs-2109-04173, DBLP:conf/aaai/Sanchez-MartinR22, DBLP:conf/nips/LouizosSMSZW17, DBLP:journals/corr/abs-2109-15062, DBLP:journals/corr/abs-2202-02195} \\
        \cline{2-5}
                                                                    & Time-Series   & Learning          & Transformers (\ref{sec:tf_inference_time}) & 
        \cite{DBLP:conf/icml/MelnychukFF22} \\
                                                                    &               &                   & VARs (\ref{sec:vars_time}) & \cite{DBLP:journals/widm/NogueiraPRPG22} \\
        \hline
        \multirow{3}{.18\linewidth}{Causal Representation Learning} & Stationary    & Disent-anglement   & Features (\ref{sec:disentanglement}) & \cite{DBLP:journals/corr/abs-1812-02230, DBLP:conf/icml/KimM18, DBLP:conf/iclr/LocatelloBLRGSB19, DBLP:conf/icml/MathieuRST19, DBLP:conf/nips/ChenLGD18, DBLP:conf/iclr/HigginsMPBGBML17, DBLP:journals/corr/Whitney16, DBLP:journals/jmlr/AchilleS18, DBLP:conf/aistats/Esmaeili0JBSPBD19, DBLP:conf/iclr/0001SB18, DBLP:conf/icml/LocatelloPRSBT20, DBLP:conf/aistats/KhemakhemKMH20, DBLP:conf/clear2/LachapelleRSEPL22, DBLP:conf/nips/GreseleKSSB21} \\
                                                                    &               &                   & Domains (\ref{sec:disentanglement}) &
        \cite{DBLP:conf/cvpr/HadadWS18, DBLP:conf/iclr/BauZSZTFT19, DBLP:conf/icml/PengHSS19, DBLP:conf/nips/GabbayCH21, DBLP:journals/corr/abs-2111-13839} \\
                                                                    &               & Augmentation & (\ref{sec:data_augmentation}) &   
        \cite{DBLP:conf/icml/GeigerWLRKIGP22, DBLP:conf/iclr/ZhangCDL18, DBLP:conf/icml/VermaLBNMLB19, DBLP:conf/wacv/Mangla0SKBK20, DBLP:journals/corr/abs-2202-08325} \\
        \hline
    \end{tabularx}
    \caption{Causality taxonomy for the listed papers. }
    \label{tab:summary-table}
\end{table}

%% file: causality_theory.tex
\section{Causality Theory}
\label{sec:causality_theory}

Causality is the field of research aiming to find systematic methods for uncovering cause-effect relationships. It aims to answer the following questions:  what evidence is needed to infer a causal link, and what can we infer from it? \cite{pearl-2009}. Causality has been studied in several domains of science and philosophy \cite{woodward2005making}, but we will restrict ourselves to the approaches from computer science.

The recovery of causal relationships can be divided into two tasks: \textit{causal structure recovery} and \textit{causal inference}. Causal structure recovery attempts to extract and represent the causal structure of the information from an unstructured source (e.g. tabular data). This task is so far exclusively 
tackled by graphical models like the Structural Causal Model (SCM) \cite{pearl-2009}.  
In contrast, causal inference aims to answer causal queries, i.e. estimate the values of causal variables in given situations. 
Causal inference methods either attempt to solve specific cases (e.g. answer a single query) or generate functions that can solve all queries (e.g. estimate the mapping functions of an SCM as explained in the next section). The term causal inference has often been used to encompass both tasks \cite{pearl-2009} but we will distinguish them throughout this survey to provide more clarity on the tasks each presented method aims to solve. The majority of the literature on Causality tackles causal inference. 

In this survey, we will focus on three paradigms for representing cause-effects relationships: Pearl's \textit{Structural Causal Model} \cite{pearl-2009}, \textit{Rubin Causal Model} \cite{rubin1974estimating}, and \textit{Granger Causality} \cite{10.5555/781840.781842}. They are discussed in Sections \ref{sec:scm}, \ref{sec:pof}, and \ref{sec:time-series} respectively.

\subsection{Structural Causal Model}
\label{sec:scm}

The Structural Causal Model \cite{pearl-2009, pearl-deductive-2014, DBLP:conf/nips/XiaLBB21} is a graphical model for representing cause-effect relationships.
A graph $\mathcal{G} = (\mathbf{V}, \mathbf{E})$ is a set of nodes (or \textit{vertices}) $\mathbf{V}$ and edges $\mathbf{E} \subset \mathbf{V} \times \mathbf{V}$ connecting the nodes, with $N=|\mathbf{V}|$ the size of the set of nodes and $M=|\mathbf{E}|$ the size of the set of edges. Edges can either be directed or undirected. In an SCM, the edges are directed to represent causation and written with an arrow $\rightarrow$.
A \textbf{directed path} in a graph is a sequence of nodes linked together by directed edges (following the direction of the arrow). A \textbf{cycle} in a graph is a directed (non-empty) path where the first and last nodes are identical. A graph containing only directed edges and without cycles is called a Directed Acyclic Graph (DAG). \textit{Parent} nodes are linked to \textit{children} nodes via an edge: $parent \rightarrow children$. The \textit{ancestors} of a node $V$ are the nodes that are on a directed path to $V$ and the \textit{descendants} of $V$ are on a directed path from $V$. The respective parent, children, ancestor and descendent sets of a node $V$ are noted  $\mathbf{pa}$, $\mathbf{ch}$,  $\mathbf{anc}$ and  $\mathbf{desc}$.
An SCM is a model based on a DAG extended with a set of functions and a probability distribution, and where the nodes are causal variables and the edges are causal dependencies between the variables. 
Each node is associated with a function taking its parents (causes) as inputs and computing an (effect) value for the node.

\subsubsection{Formal definition}
\label{sec:scm_def}

An SCM is defined as a tuple $\mathcal{M} = \langle \mathbf{U}, \mathbf{V}, \mathcal{F}, P(\mathbf{U}) \rangle$ where:
\begin{itemize}
    \item $\mathbf{U}$ is the set of unobserved or \textit{exogenous} variables
    \item $\mathbf{V}$ is the set of observed or \textit{endogenous} variables
    \item $\mathcal{F}$ is a set of mapping functions $\{ f_{V_i} \in \mathcal{F} : V_i \leftarrow f_{V_i}(\mathbf{pa}_{V_i}, \mathbf{U}_{V_i}) \}_{i \in 1 \dots |V|}$, $f_{V_i}$ computes the instance value of the endogenous variable $V_i$ given its causes: the instance values of the parent endogenous variables $\mathbf{pa}_{V_i} \subset \mathbf{V} \setminus (V_i \cup \mathbf{desc}(V_i))$ and exogenous variables $\mathbf{U}_{V_i} \subset \mathbf{U}$
    \item $P(\mathbf{U})$ is a probability function over the exogenous variables $U_i \in \mathbf{U}$
\end{itemize}

\begin{figure}
    \centering
    \begin{subfigure}{0.5\textwidth}
        \centering
        \includegraphics[scale=0.8]{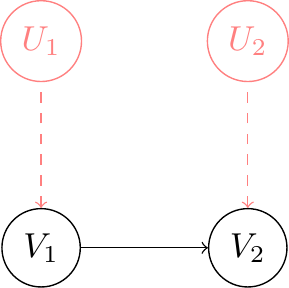}
        \Description[Structural Causal Model example]{Example of Structural Causal Model (SCM). Endogenous node V1 is connected to V2 by an edge. Exogenous node U1 is connected to V1 and, similarly, U2 is connected to V2.}
    \end{subfigure}
    \hfill
    \begin{subfigure}{0.45\textwidth}
        \centering
        $\mathbf{U} = \{U_1, U_2\}$ \\
        $\mathbf{V} = \{V_1, V_2\}$
        \begin{equation*}
          \mathcal{F} = 
            \begin{cases}
              V_1 \leftarrow f_{V_1}(U_1)\\
              V_2 \leftarrow f_{V_2}(V_1, U_2)
            \end{cases}       
        \end{equation*}
        $P(U_i) \sim \mathcal{N}(0,1)~\forall U_i \in \mathbf{U}$ \\
    \end{subfigure}
    \caption{Example of Structural Causal Model (SCM) with two endogenous variables $V_1$ and $V_2$ and two exogenous variables $U_1$ and $U_2$. $\mathcal{F}$ is the set of mapping functions linking $V_1$, $V_2$ and $U_1$, $U_2$ together. $P(\mathbf{U})$ is the probability function defined over $U_1$ and $U_2$. }
    \label{fig:scm_ex}
\end{figure}

The structure of an SCM, (i.e. the dependencies of each observed variable) can be represented with a DAG. Figure~\ref{fig:scm_ex} provides an example of an SCM. An SCM aims to model the causal mechanisms underlying the observed variables $\mathbf{V}$. The mechanisms of the unobserved variables $\mathbf{U}$ cannot be determined. The consequences on the SCM are that variables of $\mathbf{U}$ have no parents and are modelled by a probability distribution to represent uncertainty instead of a deterministic function.
The acyclicity assumption of the causal graph is useful to disambiguate the causes from the effects, as we detail in Section \ref{sec:causal_structure_discovery}.
A complete SCM can fully represent a causal system. However, building the SCM of a set of variables is a challenging task requiring the identification and estimation of the causal links between the variables (structure discovery and inference). Methods for generating an SCM are introduced in Sections \ref{sec:causal_structure_discovery} and \ref{sec:causal_inference}.

\subsubsection{Hidden confounders and Markov-equivalence}
\label{sec:confounders}

Recovering the causal structure of a system of variables is challenging and not always possible. First, the main challenge in causal tasks is dealing with confounding effects. A \textit{hidden confounder} is an exogenous variable that is a shared cause of several endogenous variables. 
In \textit{Markovian} graphs, the value of an observed variable does not depend on other observed variables than its parents (i.e. $V \independent \mathbf{anc}_V \setminus \mathbf{pa}_V |\mathbf{pa}_V$). Graphs with hidden confounders are \textit{non-Markovian} (also called \textit{semi-Markovian} if they are acyclic) because the exogenous factors $\mathbf{U}$ are not independent and can affect multiple variables $\mathbf{V}$ distant in the graph, creating non-Markovian dependencies \cite{DBLP:conf/nips/XiaLBB21}. Second, even without hidden confounders, the \textit{true} causal structure may only be recoverable up to a \textit{Markov-equivalent} class. Markov-equivalent SCMs have a different causal structure that yields the same observations.

\begin{figure}
    \begin{subfigure}[t]{0.24\textwidth}
        \centering
        \includegraphics[scale=0.9]{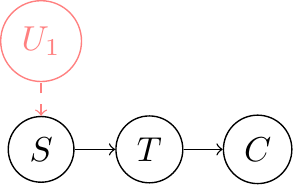}
        \Description[Markov relative DAG example]{Example of Markov-relative DAG. S is connected to T and T is connected to C. One exogenous node U1 is connected to S.}
        \caption{Markov-relative DAG. }
        \label{fig:mr_dag_eq_1}
    \end{subfigure}
    \begin{subfigure}[t]{0.24\textwidth}
        \centering
        \includegraphics[scale=0.9]{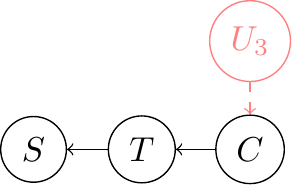}
        \Description[Markov equivalent DAG example]{Example of DAG Markov-equivalent to the previous example. C is connected to T and T is connected to S. One exogenous node U1 is connected to C.}
        \caption{Markov-equivalent DAG with another cause. }
        \label{fig:mr_dag_eq_2}
    \end{subfigure}
    \begin{subfigure}[t]{0.24\textwidth}
        \centering
        \includegraphics[scale=0.9]{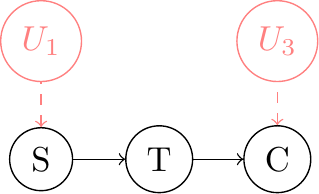}
        \Description[Markov relative DAG example with exogenous variables]{Example of Markov-relative DAG with exogenous variables. S is connected to T and T is connected to C. Two exogenous nodes U1 and U3 are connected to S and C, respectively.}
        \caption{Markov-relative DAG with exogenous variables. }
        \label{fig:mr_dag_exo}
    \end{subfigure}
    \begin{subfigure}[t]{0.24\textwidth}
        \centering
        \includegraphics[scale=0.9]{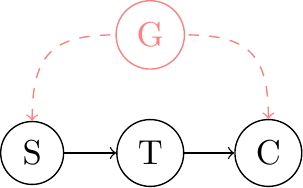}
        \Description[Semi Markov DAG example with single confounder]{Example of Semi Markov DAG with a single confounder. The graph is the same as the previous one DAG but exogenous nodes U1 and U3 have been merged to a single node G, connected to S and C, which acts as a hidden confounder if unobserved.}
        \caption{Semi-Markovian DAG with exogenous variables. }
        \label{fig:mr_dag_conf}
    \end{subfigure}
    \caption{Examples of Markov-equivalent and semi-Markovian (acyclic) DAG causal structures. (\ref{fig:mr_dag_eq_1}) a Markov-relative DAG modeling the causal effects of smoke (S) on the risk of cancer (C) with an intermediate variable corresponding to the presence of tar (T) in the lungs. (\ref{fig:mr_dag_eq_2}) a Markov-equivalent model that can be obtained with the same observational data interpreted differently. (\ref{fig:mr_dag_exo}) another Markov-relative DAG where exogenous variables forbid the previous interpretation. (\ref{fig:mr_dag_conf}) the same model where the unobserved variables have been merged to a single latent confounder: a genetic factor \textcolor{red!50}{(G)} causing both the tendency to smoke and the high risk of cancer. The last graph is semi-Markovian. }
    \label{fig:mr_dag}
\end{figure}

Figure \ref{fig:mr_dag} illustrates these two notions: we take the example of smoke and cancer. Smoking (S) is correlated with the presence of tar in the lungs (T) and tar is correlated with a high risk of cancer (C). Figures \ref{fig:mr_dag_eq_1} and~\ref{fig:mr_dag_eq_2} show two Markov-equivalent graphs that can be obtained from the same observations. In these examples, variables are systematically correlated because they are all deterministic consequences of the same exogenous process. If another source of stochasticity is involved as in Figure \ref{fig:mr_dag_exo} (yielding a different data distribution), the causal direction may be recoverable (e.g. T always follows S but C is only sometimes observed). Another possible causal structure may involve an unobserved common cause (e.g. genetics), as seen in Figure \ref{fig:mr_dag_conf} ($C \leftarrow G \rightarrow S \rightarrow T$ is another possible graph). If we can find a single causal structure that matches the data, the graph is \textbf{identifiable}. If two causal models match the observations, they may be separated by their abilities to answer interventional and counterfactual queries, as discussed in Section \ref{sec:pch}.

\subsubsection{Pearl's Causal Hierarchy}
\label{sec:pch}

The purpose of the SCM is to provide a framework for answering causal queries. These queries can be divided into three types and ordered into a hierarchy of increasing difficulty. This hierarchy is called \textit{Ladder of Causation} or \textit{Pearl's Causal Hierarchy} (PCH) \cite{Bareinboim2022OnPH} and is shown in Table \ref{tab:pch}.

\begin{table}
    \centering
    \caption{Pearl's Causal Hierarchy. Each layer corresponds to a type of causal query with an increasing level of required data. Each layer is associated with a corresponding intuitive activity and a probabilistic formula. The formulae illustrate the simplest case for each layer, assuming the minimum number of variables required. }
    \begin{tabularx}{0.8\textwidth}{cXXXX}
      \hline
      ~ & \textbf{Layer} & \textbf{Activity} & \textbf{Formula} \\
      \hline
      {\color{ACMLightBlue}$\mathcal{L}_1$} & Associational & Seeing & $P(y|x)$ \\
      {\color{ACMBlue}$\mathcal{L}_2$} & Interventional & Doing & $P(y|do(x))$ \\
      {\color{ACMDarkBlue}$\mathcal{L}_3$} & Counterfactual & Imagining & $P(y|do(x),x',y')$ \\
      \hline
    \end{tabularx}
    \label{tab:pch}
\end{table}
 
The first layer \textcolor{ACMLightBlue}{$\mathcal{L}_1$} in the hierarchy corresponds to \textit{observational} queries, i.e. queries based on data collected passively. They are of the form: "What does \textbf{X} tell us about \textbf{Y}?", modeled by the conditional probability $P(y|x)$, which can be solved using Bayesian statistics. Machine Learning based on supervised and unsupervised methods can only answer this type of query \cite{Bareinboim2022OnPH}.

The second layer \textcolor{ACMBlue}{$\mathcal{L}_2$} in the hierarchy corresponds to \textit{interventional} queries, i.e. queries about the effect of a change/intervention on an environment. They are of the form: "What will happen to \textbf{Y} if we force \textbf{X}?". A new operator is used to model this query mathematically, the \textbf{do-operator} \cite{10.1093/biomet/82.4.669}. $\mathbf{do(X=x)}$ corresponds to the forced assignment of the variable $\mathbf{X}$ to the value $\mathbf{x}$ (even if the case is never observed). The notation $\mathbf{Y_x}$ is also often used for $\mathbf{Y}|\mathbf{do(X=x)}$. This operator can be reduced into a conditional probability using the rules of \textit{do-calculus}, introduced in Section \ref{sec:causal_inference}. Reinforcement Learning can answer queries at this layer by interacting with its environment \cite{Bareinboim2022OnPH}.

The third and last layer \textcolor{ACMDarkBlue}{$\mathcal{L}_3$} represents \textit{counterfactual} queries, i.e. questions about an imagined situation. they are of the form: "What would have happened to \textbf{Y} if \textbf{X} had happened instead of \textbf{X'} (which led to \textbf{Y'})?". This type of query is the hardest to solve in the causal hierarchy as they require reasoning on things that did not happen ("What if?" questions). However, by marginalizing over the exogenous variables, the query can be reduced into the following expression (for the discrete case, assuming the simplest case with a single exogenous variable $U$) \cite{pearl-2009,Bareinboim2022OnPH}: 

\begin{equation}
    P(y|do(x),x',y') = \sum\limits_{u \in U} P(y|do(x),u)P(u|x',y')
\end{equation}

The main idea consists in connecting the factual and counterfactual worlds using the exogenous variables as shared evidence. The computation process is divided into three steps: (1) \textit{abduction} computes $P(u|x',y')$; (2) \textit{action} performs the intervention $do(x)$ and; (3) \textit{prediction} computes the final result using the operands. 

\begin{figure}
    \begin{subfigure}{0.25\textwidth}
        \centering
        \includegraphics[scale=0.9]{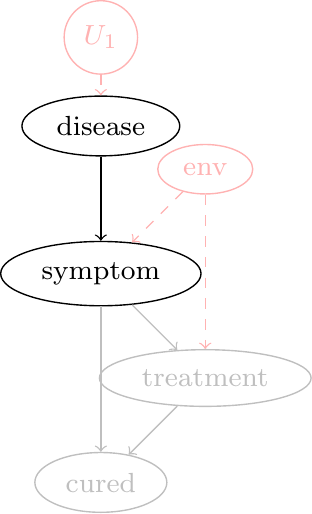}
        \caption{$\mathcal{L}_1$ query: What do the \textit{symptoms} tell us about the \textit{disease}?}
        \Description[Causal Graph Example illustrating L1 query]{Causal graph illustrating queries of type L1. It is described in the text, each variable corresponds to a node in the graph.}
    \end{subfigure}
    \hfill
    \begin{subfigure}{0.25\textwidth}
        \centering
        \includegraphics[scale=0.9]{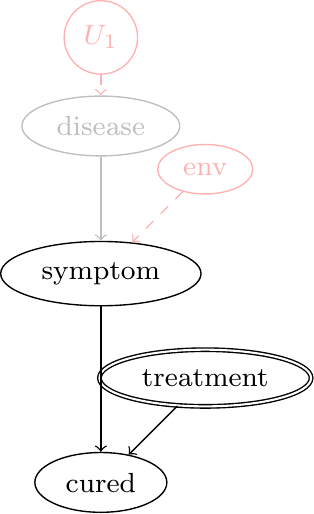}
        \caption{$\mathcal{L}_2$ query: Will the patient be \textit{cured} if he takes the \textit{treatment}?}
        \Description[Causal Graph Example illustrating L2 query]{Causal graph illustrating queries of type L2. It is the same graph as for the L1 query except the links to the treatment variable have been removed.}
    \end{subfigure}
    \hfill
    \begin{subfigure}{0.45\textwidth}
        \centering
        \includegraphics[scale=0.8]{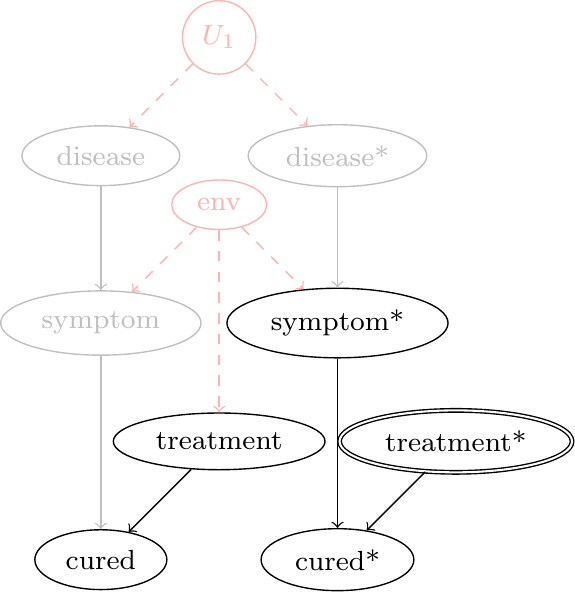}
        \caption{$\mathcal{L}_3$ query: Would the patient had been \textit{cured} if he had not taken the \textit{treatment}?}
        \Description[Causal Graph Example illustrating L3 query]{Causal graph illustrating queries of type L3. The graph from L1 and L2 queries are represented next to each other. All nodes from L1 have their L2 counterparts representing the counterfactual scenario. Only the exogenous variables U1 and environment do not change and are connected to both graph.}
    \end{subfigure}
    \caption{Causal graphs involved in the computation of layer $\mathcal{L}_i$ queries. In an $\mathcal{L}_2$ query, the intervened variable (represented with a double edge) has its value forced into the graph and does not depend on its parent variables. For the $\mathcal{L}_3$ query, the factual world and the counterfactual world (variables with a *) are linked by the same exogenous factors. The observations from the factual worlds are used to estimate the $U_1$ and $env$ variables (abduction), then the counterfactual graph is intervened on (action) and the final estimation is computed (prediction). }
    \label{fig:layer_queries}
\end{figure}

Figure \ref{fig:layer_queries} illustrates each type of query with an example. A \textit{disease} propagates in a population with a mechanism not covered in our causal model and represented by an exogenous variable $U_1$. When infected with the disease, a patient will develop \textit{symptoms}. The symptoms can be mild or severe depending on \textit{environmental factors} (e.g. healthcare access may reduce the severity of the symptoms), not observed. The patient can then take a \textit{treatment} to be \textit{cured}. The access to the treatment may depend on unobserved factors correlated with the symptoms (e.g. the treatment may be accessible with the same healthcare system). The environmental factor here acts as a hidden confounder. The problem in this situation is assessing the effect of the treatment on the probability of getting cured. In our model, the answer depends (or not) on the treatment and the severity of the symptoms, but the two variables are not independent as they share a common hidden cause.

An SCM can accurately represent causal mechanisms. All query layers in the PCH can be computed with a fully specified SCM. Under the SCM framework, causal inference aims to estimate the true causal model as closely as possible. However, recovering the true SCM is hard to achieve, and the \textbf{Causal Hierarchy Theorem} (CHT) \cite{Bareinboim2022OnPH} furthermore states that the amount of information needed to solve a query at a layer $\mathcal{L}_i$ is not sufficient in general to solve queries at layers $\mathcal{L}_{i+1}$. In particular, the CHT states that for a causal model able to solve $\mathcal{L}_1$ queries, it is always possible to find a second model that yield the same responses to $\mathcal{L}_1$ queries but different responses to \textbf{$\mathcal{L}_2$} queries. Collapses can be observed between $\mathcal{L}_2$ and $\mathcal{L}_3$ for exceptional SCMs but it is in general possible to find two causal models with the same responses to $\mathcal{L}_2$ but different responses to $\mathcal{L}_3$ queries. This information acquisition problem is the main challenge in Causality as once achieved, the problem can be reduced to probabilistic reasoning \cite{DBLP:journals/corr/abs-2111-13936}.

\subsection{Rubin Causal Model}
\label{sec:pof}

The Rubin Model, or Neyman-Rubin Causal Model, or Potential Outcome Framework (POF) \cite{splawa-neyman-application-1990, rubin1974estimating, DBLP:journals/tkdd/YaoCLLGZ21}, is a causal model based on the idea of observed and counterfactual \textit{outcomes}. As opposed to the SCM, this model does not aim to recover the causal structure of the data but to estimate the distribution of an outcome given the application of a \textit{treatment} variable.
This situation is often encountered in medical applications where researchers would like to estimate the effect of a treatment on a disease. A standard method for achieving it is to perform a \textit{Randomized Control trial} where the treatment is randomly assigned to a group of people to mitigate confounding effects, e.g. age, ethnicity, and health status of the patients. However, this is a cost-intensive method and may not be achievable in some situations for ethical, feasibility or physical reasons. For instance, injecting medication with a high death rate into a treatment group is unethical. Alternatively, when measuring the effects of an economic decision on a country, it is infeasible to have a treated and a control group. It is even impossible when working on a unique subject: after deciding whether to take an action against climate change or not, only one of the two outcomes is observed. The other, called \textit{counterfactual outcome} can never be observed.
The latter example is an illustration of the \textbf{fundamental problem of causal inference}: we would like to estimate the outcome of a situation that is not observed.

\subsubsection{Formal definition}

In the Rubin model, a \textit{treatment} is applied to a subject of interest, called \textit{unit}. The treatment variable is denoted $X \in \{0,1, \dots, N_X\}$ with $N_X$ the number of possible treatments and $0$ representing the absence of treatment.
We note the random variable corresponding to the potential outcome as $Y(X = x)$, $Y_x$ or $Y|do(x)$ to be consistent with the notation from the previous section. $Y_0$ represents the outcome for the control group and $Y_1$ the outcome for the group with treatment 1 (or the treated group if only one treatment is studied).
The background random variables, or \textit{covariates}, corresponding to the attributes of the patients are denoted $Z$. We note that alternative notations can be used in the literature, we choose this one to maintain a consistent theoretical notation with the one used by the SCM.
Within the POF, the goal of causal inference is to determine the treatment effect at different levels of granularity: population level, subgroup level, or individual level.

\paragraph{Average Treatment Effect (ATE)} The ATE, defined in Equation \ref{eq:ate}, is the expected average difference of effect on the outcome between units taking the treatment and units not taking it.

\begin{equation}
    \text{ATE} = \mathbb{E}[Y_1 - Y_0]
    \label{eq:ate}
\end{equation}

\paragraph{Average Treatment Effect on the treated group (ATT)} The ATT, defined in Equation \ref{eq:att}, is the difference between the expected outcome for the treated group and the counterfactual outcome if the same treated group did not take the treatment.

\begin{equation}
    \text{ATT} = \mathbb{E}[Y_1|X = 1] - \mathbb{E}[Y_0|X = 1]
    \label{eq:att}
\end{equation}

\paragraph{Conditional Average Treatment Effect (CATE)} The CATE, defined in Equation \ref{eq:cate}, is the difference of effect on the outcome between units taking the treatment and units not taking it, given a subgroup conditioned on a background variable $Z$.

\begin{equation}
    \text{CATE} = \mathbb{E}[Y_1|Z = z] - \mathbb{E}[Y_0|Z = z]
    \label{eq:cate}
\end{equation}

\paragraph{Individual Treatment Effect (ITE)} The ITE, defined in Equation \ref{eq:ite}, for a particular individual or unit \textit{i}, is the difference of effect on the outcome between a situation where the treatment is applied and one where the treatment is not. In this last situation, it appears that one of the situations cannot be observed as a single individual cannot take the treatment and not take it at the same time. This counterfactual outcome (which could be either $Y_1$ or $Y_0$) must be determined with other methods.

\begin{equation}
    \text{ITE}^{(i)} = Y_1^{(i)} - Y_0^{(i)}
    \label{eq:ite}
\end{equation}

The goal of causal inference with the Rubin model is to compute the above formulae, the final objective being the computation of the ITE for any given individual. In order to achieve this goal, it is necessary to compute counterfactual outcomes despite the lack of observations. Techniques have been developed to this end, detailed in Section \ref{sec:causal_inference}.

\subsection{Granger Causality} 
\label{sec:time-series}

The reader may have noticed that the notion of Causality we have mentioned so far does not include any notion of time. It may seem surprising as it is a fundamental notion in our intuitive understanding of causality and a solid inductive bias when causal inference is conducted by the brain \cite{bramley-time-2018}.
However, as we have seen in the previous sections, time is not central and not even needed in most popular causal frameworks. We are now interested in the other end of the spectrum, causal models that rely on time-sensitive information and apply causal learning to time series.

A time series is a set of variables with values evolving through time $\mathbf{X}(t) = (x_1(t), x_2(t), \dots, x_N(t))$ where $t$ corresponds to the current timestep, $\{X_i \in \mathbf{X}\}_{i \in 1 \dots N}$ is the set of $N$ time-sensitive variables, and $\mathbf{X}(t)$ is a snapshot of the values for each variable of $\mathbf{X}$ at timestep $t$.
Granger Causality \cite{10.5555/781840.781842} is a theory of Causality developed for time series. In this model, a series $\mathbf{X}$ causes (or "Granger-causes" \cite{DBLP:journals/kais/MoraffahSKBWTRL21}) $\mathbf{Y}$ if we can better predict $\mathbf{Y}(t)$ using all the information available, denoted $\mathbf{U}$, than with all the information deprived of the past values of $\mathbf{X}$: $\mathbf{U} \setminus \{\mathbf{X}(0), \mathbf{X}(1), \dots, \mathbf{X}(t-1)\}$. More formally: if $\sigma^2_t(\mathbf{Y}|\mathbf{U}) < \sigma^2_t(\mathbf{Y}|\mathbf{U} \setminus \{\mathbf{X}(0), \mathbf{X}(1), \dots, \mathbf{X}(t-1)\})$, where $\sigma^2_t(\mathbf{A}|\mathbf{B})$ is the variance of the predictive error $\epsilon_t(\mathbf{A}|\mathbf{B}) = \mathbf{A}(t) - P(\mathbf{A}(t)|\mathbf{B})$. It is important to highlight that Granger Causality does not recover true causes \cite{maziarz2015review}. In the above definition, if $\mathbf{X}$ and $\mathbf{Y}$ are correlated but caused by a third hidden series $\mathbf{Z}$ ($\mathbf{X} \leftarrow \mathbf{Z} \rightarrow \mathbf{Y}$), the Granger Causality test will not differenciate it from the causal case $\mathbf{X} \rightarrow \mathbf{Y}$.

Another line of methods for inferring causality from time series is based on Information Theory and uses entropy measures to infer causal relationships. The Shannon entropy \cite{DBLP:journals/bstj/Shannon48} of a variable $X$ can defined as shown in Equation \ref{eq:entropy}. similarly, we can define the \textit{joint entropy} between two variables $X$ and $Y$, shown in Equation \ref{eq:joint_entropy}, and the \textit{conditional entropy} of $X$ given $Z$, shown in Equation \ref{eq:conditional_entropy}.

\begin{align}
    \label{eq:entropy}
    H(X) &= -\sum\limits_x P(x) \log P(x)\\
    \label{eq:joint_entropy}
    H(X, Y) &= -\sum\limits_x \sum\limits_y P(x, y) \log P(x, y)\\
    \label{eq:conditional_entropy}
    H(X|Z) &= -\sum\limits_x \sum\limits_z P(x, z) \log P(x|z)
\end{align}

Entropy measures the amount of uncertainty from a variable. The lower the entropy, the easier the value of a random variable can be predicted. From these definitions, we can define the \textbf{Mutual Information} between two variables $X$ and $Y$, given $Z$:

\begin{equation}
    I(X; Y|Z) = H(X|Z) + H(Y[Z) - H(X, Y|Z)
    \label{eq:mutual_shared_info}
\end{equation}

The Mutual Information between two variables represents the amount of information shared between the two. Methods based on Mutual Information attempt to detect the flow of information in the causal system over time \cite{palus2001}. The above formulae are symmetric and therefore do not account for the direction taken by the information. In order to alleviate this problem, the concept of \textbf{Transfer Entropy} was introduced by \citet{PhysRevLett.85.461} to detect asymmetries between variables from the data. Transfer Entropy is defined in Equation \ref{eq:transfer_entropy} (timesteps are subscript for simplicity).

\begin{equation}
    T_{X \rightarrow Y} = H(Y_t|Y_{t-1}, Y_{t-2}, \dots, Y_{t-K}) - H(Y_t|Y_{t-1}, \dots, Y_{t-K}, X_{t-1}, \dots, X_{t-K})
    \label{eq:transfer_entropy}
\end{equation}

Transfer Entropy is an asymmetric measure of the information flow from $X$ to $Y$. It compares the entropy of process $Y$ at timestep $t$, given its past values up to step $t-K$, and the entropy of $Y$ given its past values and the ones of a second variable $X$. It can be used to recover the causal structure between time series. Granger Causality and Transfer Entropy are two interlinked concepts, particularly in the case of Gaussian variables where the two measures are equivalent \cite{PhysRevLett.103.238701}.
Transfer Entropy works well with two processes but has trouble when multiple variables can influence the system \cite{DBLP:journals/kais/MoraffahSKBWTRL21, DBLP:journals/siamads/0007TB15}. Further work attempt to overcome this issue by proposing new measures for causality in time series \cite{VAKORIN2009152, PhysRevE.82.016207, DBLP:journals/siamads/0007TB15}.

%% file: neural_causal_model.tex
\section{Neural Causal Model}
\label{sec:neural_causal_model}

Causality offers a large set of tools for reasoning over causes and effects, but despite theoretical correctness, these traditional approaches struggle when dealing with high-dimensional variables. Conversely, Deep Learning has been proven very successful at extracting patterns in large datasets and applying them to various tasks. However, they do not generalize well out of their initial distribution, show poor causal reasoning capabilities, and can easily be fooled by confounding effects \cite{DBLP:journals/corr/GoodfellowSS14, DBLP:conf/icml/KohSMXZBHYPGLDS21, DBLP:journals/corr/abs-2011-15091, DBLP:journals/corr/abs-2204-00607}. It seems natural to combine the best of both worlds, but interestingly, the two fields have arisen in separate ways until very recently. Attempts have been made to generate neural models capable of causal reasoning, but these approaches are still stammering and how to develop causal reasoning engines at scale remains an open problem.
This section focuses on the potential benefits of causal reasoning engines and the theory that has been developed. Sections \ref{sec:causal_structure_discovery} and \ref{sec:causal_inference} will introduce causal structure discovery and inference methods based on the framework along methods based on the farameworks from Section \ref{sec:causality_theory}. Section \ref{sec:representation_learning} will address the problem of representation learning under a causal perspective, i.e. the problem of learning causal variables from low-level data.

\subsection{Motivations}

As mentioned in Section \ref{sec:pch}, traditional Machine Learning methods based on supervised learning rely only on observations. The CHT \cite{Bareinboim2022OnPH} states that such models cannot answer causal queries in general (queries in the second and third layers in the hierarchy of Table \ref{tab:pch}). Adding Causality theory into Machine Learning techniques, and Deep Learning (DL) in particular, could help leverage this issue and bring additional improvements to the field. We summarize the potential impact causal models can bring to Deep Learning.

\subsubsection{Generalization}

First, modern supervised learning techniques rely on the assumption that the data are \textit{independent and identically distributed} (i.i.d). This assumption can lead to brittle models \cite{DBLP:journals/corr/GoodfellowSS14, DBLP:conf/icml/KohSMXZBHYPGLDS21, DBLP:journals/corr/abs-2011-15091, DBLP:journals/corr/abs-2204-00607} in environments where it does not hold. 
The i.i.d assumption ensures that the learned patterns can be applied in general cases. However, datasets can be biased or not representative enough of real-world distributions. One model may perform well on one image dataset but poorly on another one where the distribution is slightly different (e.g. a model trained on photographs may not perform well on sketches). Models are not only prone to bias, but can also be fooled by spurious examples, adversarial attacks and distribution shifts \cite{DBLP:journals/corr/abs-2011-15091, DBLP:journals/pieee/ScholkopfLBKKGB21, DBLP:journals/corr/abs-2204-00607}. To counter this effect, many techniques have been attempted, the most successful consisting in increasing as much as possible the size of the model and the data, leading to the \textbf{data-centered} approach to DL, dominated by the \textit{Foundation models} \cite{DBLP:journals/corr/abs-2108-07258}.
Causal models, on the other hand, can be considered \textbf{model-centered} approaches as they do not rely on the absence of spurious effects in the data. Causal models aim to represent the causal \textit{structure} of data-generating mechanisms, invariant to the way the data is distributed. 
Moreover, they acknowledge the spurious correlations and the confounding effects and can take them into account, as seen in Section \ref{sec:causality_theory}, without needing much data. Causal models have recently proven to less overfit and yield better, faster, and more robust results in low-data regimes \cite{DBLP:journals/corr/abs-2206-04620}. Incorporating causal engines into DL tasks can lead to more robust models, less prone to bias, and with better generalization.

\subsubsection{Modularity}

Second, DL engines are usually black boxes made of dense and entangled computations. It makes them hardly interpretable and very computation-intensive. A great line of work argues that modular architectures would help tackle those two downsides and learn transferable mechanisms that could be reused from one task to the next \cite{goyal-recurrent-2020, rahaman-dynamic-2021, DBLP:journals/corr/abs-2108-07258}. 
In particular, the \textbf{Independent Causal Mechanism} (ICM) principle \cite{10.5555/3202377, DBLP:journals/pieee/ScholkopfLBKKGB21} is a postulate stating that causal mechanisms are autonomous and do not interact with each other. The \textbf{Sparse Mechanism Shift} (SMS) hypothesis \cite{DBLP:journals/pieee/ScholkopfLBKKGB21} furthermore adds that small changes in the distribution generate only local changes in a causal model, not affecting all factors. Dense computations should not be necessary and only yield additional spurious correlations. Recent work showed that modular neural networks have a much lower gradient magnitude and work better on tasks with sparse causal structure \cite{DBLP:journals/corr/abs-2206-04620}.
Within the DL field, several attempts have been made to generate such sparse and modular architectures \cite{goyal-recurrent-2020, fedus-switch-2021, du-glam-2021, rahaman-dynamic-2021}. They mostly rely on \textit{Mixtures-of-Experts} and use routing mechanisms for selecting the right experts. These architectures are composed of stacks of experts and do not have a graph structure. Their modularity is still quite limited. Graphical causal models, on the other hand, are inherently based on a graph structure, usually sparse, that would be very appropriate to this end.
Theories from neuroscience also posit that the brain works in a modular way. Each module is an expert in a domain and communicates sparsely with others experts, transmitting only few information \cite{Baars1997INTT, doi:10.1080/00131911.2021.1930914}. Causal models composed of a DAG with nodes sparsely exchanging information would be closer to the actual behavior of the brain \cite{DBLP:journals/corr/abs-2011-15091}.

\subsubsection{Interventions and Counterfactuals}

Third, as mentioned at the beginning of this section, DL models currently do not have the ability to answer queries of the type "what would happen if we do X? What if Y had happened instead?" which correspond to causal queries of layer $\mathcal{L}_2$ and $\mathcal{L}_3$, as specified by the CHT in Section \ref{sec:pch}. Online \textit{Reinforcement Learning} (RL) techniques can interact with the environment and by this mean directly perform interventions. Therefore, they can answer some layer $\mathcal{L}_2$ queries but causal reasoning may help improve the capabilities of RL models at reasoning in an imagined space and answer counterfactuals queries \cite{DBLP:journals/corr/abs-1901-08162, DBLP:journals/pieee/ScholkopfLBKKGB21}.
Going further, we can state that counterfactual reasoning is a critical key towards more human-like AI agents. An essential step towards this goal is believed to be the generation of \textbf{World Models}, i.e. models able to understand the inner mechanisms of their surroundings and perform reasoning and planning in their environment \cite{DBLP:journals/corr/Schmidhuber15, DBLP:journals/corr/abs-1803-10122, DBLP:journals/nn/MatsuoLSPSSUM22}.

\subsection{The Neural Causal Model}

The \textit{Neural Causal Model} (NCM) \cite{DBLP:conf/nips/XiaLBB21} is a theoretical framework providing a link between SCMs and neural networks.
An NCM is defined, similarly as the SCM defined in Section \ref{sec:scm_def}, as a triple $\mathcal{M}(\boldsymbol{
\theta
}) = \langle \mathbf{U}, \mathbf{V}, \mathcal{F}, P(\mathbf{U}) \rangle$, where:

\begin{itemize}
    \item as for the SCM, $\mathbf{U}$ is the set of exogenous variables
    \item as for the SCM, $\mathbf{V}$ is the set of endogenous variables
    \item as for the SCM, $\mathcal{F}$ is a set of mapping functions computing the effect value of a variable given its causes (instance values of its parent endogenous variables $\mathbf{pa}_{V_i} \subset \mathbf{V} \setminus (V_i \cup \mathbf{desc}(V_i))$ and exogenous variables $\mathbf{U}_{V_i} \subset \mathbf{U}$), the NCM restricts the class of functions to 
    \textit{feedforward neural networks}:
    $\{ f_i \in \mathcal{F} : v_i \leftarrow f_{V_i}(
    \theta_{V_i}; \mathbf{pa}_{V_i}, \mathbf{U}_{V_i}) \}_{i \in 1 \dots |V|}$, $f_{V_i}$ 
    \item $\boldsymbol{\theta} = \{\theta_{V_i} : V_i \in \mathbf{V} \}$
    is the set of parameters for the mapping functions
    \item as for the SCM $P(\mathbf{U})$ is the probability function defined over the noise variables in $U_i \in \mathbf{U}$, the NCM furthermore constraints the probability law to 
    $\mathbf{u_i} \sim \text{Unif}(0,1) ~\forall U_i \in \mathbf{U}$
\end{itemize}

In a nutshell, NCMs are a subclass of SCMs where the mapping functions representing the causal relationships between the variables are limited to the class of parametrized neural networks. The initial paper \cite{DBLP:conf/nips/XiaLBB21} furthermore restrict the distribution of exogenous variables to a uniform law, and the mapping functions to Multi-Layer Perceptrons (MLPs). As MLPs are universal function approximators and all distributions can be estimated from the uniform law \cite{angus1994probability}, this restriction guarantees that the NCM class has the same expressivity as the broader SCM class. NCMs are able to represent queries at layers $\mathcal{L}_1$, $\mathcal{L}_2$ and layers $\mathcal{L}_3$ (observations, interventions and counterfactual) \cite{DBLP:conf/nips/XiaLBB21}. In practice, however, causal inference is intractable with MLPs \cite{DBLP:journals/corr/abs-2110-12052}. Subsequent works relax these two assumptions. This choice can leads to a loss in expressivity but yields learnable models. We also emphasize that works combining neural models with causal engines do not necessarily follow the NCM principles.

%% file: causal_discovery.tex
\section{Causal Structure Discovery}
\label{sec:causal_structure_discovery}

Causal structure discovery is the task of learning the causal relationships in a set of variables, whether or not two variables are causally linked, and the direction of the causation when this is possible. 
The structure discovery methods can be divided into multiple categories: 
Constraint-based, score-based, and non-Gaussian methods are the traditional approaches to causal structure discovery. Some non-Gaussian methods also take advantage of neural models. In constrast, Variational and Transformer approaches rely on recent Deep Learning techniques.

\subsection{Constraint-based methods}
\label{sec:constraints}

The \textit{Constraint-based} causal structure discovery methods rely on independence tests to create causal connections between variables \cite{DBLP:journals/widm/NogueiraPRPG22} and build an SCM. They are based on the Inductive Causation (IC) algorithm \cite{DBLP:conf/uai/VermaP90}. In the original article, this algorithm can be used in the presence of hidden confounders. However, following the separation made in \citet{pearl-2009} and in order to help comprehension, we first introduce a version of IC in an unconfoundedness regime, i.e. we assume that there are no hidden confounders in the model.

The IC algorithm has three steps. The first step builds undirected causal dependencies. An undirected graph is constructed from the set of (observed) variables of interest \textbf{V}. Edges between two variables $A$ and $B$ are added to the graph if it is not possible to find a set $S_{AB} \subset V \times V \setminus {A,B}$ s.t. $A \independent B | S_{AB}$. In other words, $A$ and $B$ share a dependency that does not depend on other variables. This condition is considered sufficient to state that $A$ and $B$ are causally linked.
The second and third steps find the direction of causality. 
The second step creates $\nu\text{-}structures$, i.e. structures with the shape $A \rightarrow C \leftarrow B$ (also called colliders). For each pair of nonadjacent variables $(A, B)$ which have a neighbor $C$ in common ($A-C-B$), if $C \notin S_{AB}$, it can be deduced that $C$ does not have a causal effect on $A$ and $B$ and cannot be a cause for one of them. The three variables can, therefore, only have a $A \rightarrow C \leftarrow B$ structure. 
The third step orients the remaining possible edges. If orienting an edge in one direction creates a cycle, the edge has to be oriented in the other direction. The process is also applied if it creates a $\nu\text{-}structure$, as all structures should be discovered in the second step.

The steps of the IC algorithm are abstract and do not offer a direct implementation. Many algorithms will build upon the ideas developed in IC and provide practical implementation of the method. A first application is the SGS algorithm, which has exponential complexity in the worst case \cite{spirtes1989causality}. One of the most popular is the Peters and Clark (PC) algorithm \cite{DBLP:books/daglib/0023012}, which runs in polynomial time. Improvements attempt to reduce the complexity and increase the stability of PC by reducing the number of conditional independence tests and relaxing assumptions on the order of nodes, like PC-stable \cite{DBLP:journals/jmlr/ColomboM14}, conservative-PC \cite{DBLP:conf/uai/RamseyZS06}, PC-select (called PC-simple in the original paper) \cite{buhlmann2010variable}, or MMHC \cite{DBLP:journals/ml/TsamardinosBA06}, a PC-based method guided with a min-max heuristic. All the above methods assume that no hidden confounders are hidden in the model. This assumption helps simplify the problem but is rarely true in practice. 
\newline

We now introduce methods for recovering the graph structure in the presence of hidden confounders. If the performed tests are statistically significant, algorithms in the unconfoundedness regime can recover the causal DAG structure up to a Markov-equivalence class \cite{DBLP:conf/uai/Meek95, DBLP:conf/uai/VermaP90, Flesch2007}. This is no longer the case in the confoundness regime as there may not be a DAG structure that can account for the observed data. The IC algorithm with hidden confounders (called IC* in \citet{pearl-2009}) has its third step modified to account for confounding effects: edges are left undirected when the causal relationship $A-B$ cannot be disambiguated between $A \rightarrow B$, $B \rightarrow A$ and $A \leftarrow Z \rightarrow B$, with $Z$ a hidden confounder.

The Fast Causal Inference (FCI) algorithm  \cite{DBLP:books/daglib/0023012} provides a way to implement IC in the presence of hidden confounders, and the Real Fast Causal Inference (RFCI) algorithm \cite{colombo-learning-2012} reduces its running time. More recently, the Iterative Causal Discovery (ICD) method \cite{DBLP:conf/nips/RohekarNGN21} reduces the number of conditional tests needed by working iteratively. The algorithm starts with a full graph and progressively removes wrong edges. Independence tests are conditioned on a local neighborhood, with the size of this neighborhood increasing at each step, but as the graph becomes sparser at each step with the removal of the edges, the causal inference remains fast even with an extensive set of variables in the neighborhood. 

Finally, these approaches have also been adapted to time series. The Time series Fast Causal Inference (TsFCI) algorithm \cite{entner20120} is an adaptation of the FCI algorithm to time series. The method uses a sliding window to divide the series into several data frames, each with its own set of variables, where the variables are treated without their time component. Steps from the same time series are separate variables within a frame, so additional rules are added to consider this background knowledge.
More recently, PCMCI \cite{doi:10.1126/sciadv.aau4996} combines the PC algorithm with Momentary Conditional Independence testing. The framework is divided into two stages: a first stage removes parent candidates for each variable until a minimal set is reached, and a second stage creates the causal links from this small amount of variables.

\subsection{Score-based methods}
\label{sec:scores}

The score-based methods compute a likelihood score for each potential causal graph given a set of variables. As the size of the search space of possible graphs increases exponentially with the number of nodes, the main challenge of this class of algorithms is to prune the search space \cite{DBLP:journals/widm/NogueiraPRPG22}. The joint probability distribution over the variables \textbf{V} with a causal structure given by a graph $\mathcal{G}$ can be given by a Bayesian network $\mathcal{B}$, as shown in Equation \ref{eq:bayes_dag}. The methods described in this section attempt to recover the true DAG $\mathcal{G}$ matching the distribution.

\begin{equation}
    P_{\mathcal{B}}(V_1=v_1, V_2=v_2, \dots, V_N=v_N) = \prod\limits_{i=1}^N P(V_i=v_i|\mathbf{pa}_{V_i}^{\mathcal{G}})
    \label{eq:bayes_dag}
\end{equation}

An early search method is the K2 procedure \cite{DBLP:journals/ml/CooperH92}. It is based on a greedy search on parent nodes. Parents are added to a node if adding them increases the probability of the structure. However, the scoring done by K2 is not strictly equivalent to conditional independence as they assume priors on the data distribution \cite{DBLP:journals/jmlr/Chickering02a}. The popular Greedy Equivalence Search (GES) method \cite{meek1997graphical, DBLP:journals/jmlr/Chickering02a} iteratively constructs the causal graph from an empty DAG in two stages: in the first stage, an edge is added at each step if it increases the score of the DAG. The second stage starts when the score no longer increases. An edge is removed at each step if it increases the score of the DAG. Like the constraint-based methods, this method is exact and returns the optimal DAG (up to an equivalent class, assuming no confounders). The Greedy Interventional Equivalence Search (GIES) \cite{DBLP:conf/uai/HauserB11} improves this method by performing interventions, restricting the equivalent solution class to a smaller subset that remains valid when performing interventions. The Fast Greedy Equivalence Search (fGES) \cite{ramsey2017million} modifies the GES algorithm to allow parallelization and scaling of the algorithm to a high number of variables. These methods are an efficient way to compute a causal model, but they all assume no hidden confounders, the Greedy Fast Causal Inference (GCFI) removes this restriction by combining GES with FCI \cite{DBLP:conf/pgm/OgarrioSR16}.
In constrast with the previous works which evolve in the discrete domain, the Non-combinatorial Optimization via Trace Exponential and Augmented lagRangian for Structure learning (NOTEARS) model \cite{DBLP:conf/nips/ZhengARX18} considers DAG learning as a continuous optimization problem. The node values are represented using a feature matrix $\mathbf{X}$ and the edges of the graph are represented with a continuous weighted adjacency matrix $\mathbf{W}$. The method further assumes that the relationships between the causal variables are linear. Following this assumption, the loss function to minimize can be simply defined as $\mathcal{L}(\mathbf{W}) = \lVert \mathbf{X} - \mathbf{X}\mathbf{W} \rVert_2$. A regularization term enhances the sparsity of the graph and a constraint ensures the graph is acyclic (and nontrivial). the discrete adjacency matrix can be obtained by considering an edge exists if its corresponding weight in $\mathbf{W}$ is non-zero. This approach alleviates the issue of the combinatorial search space of DAGs. However, \citet{DBLP:journals/npl/KaiserS22} pointed out that NOTEARS leads to biased graphs when the causal variables have different variances and argue that the method is not suited for causal discovery.

\subsection{Non-Gaussian methods}
\label{sec:non_gaussian}

Constraint-based and score-based methods are the fundamental approaches to causal discovery. With the exception of NOTEARS, all rely on statistical tests. However, more recently, new directions have been studied to tackle the problem without the need for statistical tests, which are computation-intensive and require a lot of data to be statistically significant \cite{10.3389/fgene.2019.00524}. The proposed methods take inspiration from the field of Independent Component Analysis (ICA) \cite{DBLP:journals/nn/HyvarinenO00} and rely on the assumption that the data follows a non-Gaussian distribution.
This assumption allows extracting \textit{asymmetry} in the distribution of causes and effects in the data and determining the causal structure. These methods furthermore assume that no confounders are present in the model.

A first work, the Linear Non-Gaussian Acyclic Model (LiNGAM) \cite{DBLP:journals/jmlr/ShimizuHHK06} represents the causal links between variables as linear functions with the form:

\begin{equation}
    v_i = \sum\limits_{v_j \in \mathbf{pa}_{v_i}} b_{ij}v_j + \epsilon_i + c_i
\end{equation}

\noindent where any variable in the model is a linear function of its parents. $\epsilon_i$ is a disturbance term, and $c_i$ is an additional constant. The disturbance term accounts for noise (or exogenous variables in the SCM), and all $\epsilon_i$ are assumed independent (no confounders assumption). The second and most crucial assumption in LiNGAM is that the distribution of the $\epsilon_i$ is non-Gaussian. 
This restriction comes from the field of Independent Component Analysis (ICA) where this assumption is used to separate linear combinations of independent variables. If the mixture follows a Gaussian distribution, it is not possible to recover the variables that compose them (a consequence of the central limit theorem). If the mixture and all the underlying variables are non-Gaussian, it becomes possible to use their asymmetries to tear them apart. 
Going back to our problem, in matrix form, the above equation can be written as:

\begin{align}
    \mathbf{v} &= \mathbf{Bv} + \mathbf{\epsilon}\\
    \mathbf{v} &= (1-\mathbf{B})^{-1}\mathbf{\epsilon}\\
    \mathbf{v} &= \mathbf{A\epsilon}
\end{align}

The problem to solve can then be reduced to computing the values of the matrix $\mathbf{A}$, which is achieved in LiNGAM using ICA: all mixtures $v \in \mathbf{v}$ are decomposed into their independent (cause) variables from $\mathbf{v} \setminus v$.
A direct extension of LiNGAM is the Post-Nonlinear model (PNL) \cite{zhang2010distinguishing, DBLP:conf/uai/ZhangH09} which replaces the linear mapping with a composition of nonlinear functions: $v_i = f_2(f_1(v_j) + \epsilon_i)$.

Traditional methods for causal structure discovery and causal inference work well when the number of dimensions is low and sufficient data is provided to apply statistical tests. However, despite the improvements in causal structure discovery, it remains hard to recover a causal structure at scale \cite{DBLP:journals/pieee/ScholkopfLBKKGB21}. To alleviate this issue, we investigate neural methods for structure discovery.
Despite its name, the \textbf{Neural Causal Inference Network} (NCINet) \cite{DBLP:journals/corr/abs-2204-00762} is a causal structure discovery framework for two variables settings based on neural networks. The resulting graph can therefore be \textit{causal} ($X \rightarrow Y$), \textit{anticausal} ($X \leftarrow Y$), or \textit{unassociated} ($X~Y$). The core idea is that the mean squared error of a regression in the causal direction should be lower than a regression in the anticausal direction, i.e. if $X$ causes $Y$ then $y = f_1(x)$ should have a better predictive power than $x = f_2(y)$. A neural encoder generates hidden representations $\mathbf{H}_X$ and $\mathbf{H}_Y$ for variables $X$ and $Y$ and feeds them to two regressors, attempting to learn functions $f_1$ and $f_2$ respectively. The result is given to a classifier that outputs the type of relation between the two variables (causal, anticausal or unassociated). 

\subsection{Variational Models}
\label{sec:variational_discovery}

The\textbf{ Variational Causal Network} (VCN) \cite{DBLP:journals/corr/abs-2106-07635} is a method for learning the adjacency matrix of a causal graph iteratively from data. The graph is constructed autoregressively by recurrently feeding it to a Long-Short Term Memory (LSTM) network $q_\theta$, parametrized by $\theta$, that generates the graph $\mathcal{G}$.
The model maximizes the following \textit{Evidence Lower Bound} (ELBO) loss:

\begin{equation}
    \mathcal{L} = \mathbb{E}_{q_\theta(\mathcal{G})} [\log P(\mathcal{D}|\mathcal{G})] - \text{KL}(q_\theta(\mathcal{G})||P(\mathcal{G}))
\end{equation}

\noindent where $\mathcal{D}$ is the data distribution, and $P(\mathcal{D}|\mathcal{G})$ represents the probability of obtaining the data given the causal graph. The first term of the loss verifies that $\mathcal{G}$ can yield the data distribution $\mathcal{D}$. $P(\mathcal{G})$ is the prior distribution of the generated causal graph, i.e. the prior knowledge about the causal graph (here, acyclic graph). The second term uses Kullback-Leibler divergence to ensure that the generated graph follows the prior distribution.  
The VCN uses a modified version of the ELBO loss used by variational models \cite{DBLP:journals/corr/KingmaW13}. The latter are also used for causal inference (Section \ref{sec:vgae}) and causal representation learning (Section \ref{sec:disentanglement}).

\subsection{Large Language Models}
\label{sec:llm_discovery}

A recent body of work studied the capabilities of large Language Models (LLMs) at performing causal structure discovery. The studies focus on \textit{in-context learning} \cite{DBLP:conf/iclr/XieRL022}, i.e. they prompt the models to answer questions from a short description of the problem and a few example. No explicit training is performed on the given task.
\citet{DBLP:journals/corr/abs-2308-13067} performed experiments on multiple language models, including the popular GPT-3 \cite{DBLP:conf/nips/BrownMRSKDNSSAA20}. When prompted to tell if a causal link exists between two concepts, the LLM responses were mostly accurate. However, they can change significantly depending on the wording used in the prompt, with knowledge being inconsistent. While these experiments showed that current LLMs do not exhibit robust causal reasoning capabilities, they also highlight the difficulty to assess if a model understand the causal links or recite them. In particular, the authors argue that LLMs can exhibit good performance on $\mathcal{L}_2$ and $\mathcal{L}_3$ queries of the CHT by learning the structure of the true causal graph in their training data.
To assess if LLMs can discover new causal structures (not leaked in the training data), \citet{DBLP:journals/corr/abs-2306-05836} proposes an abstract causal discovery evaluation benchmark. The authors show that LLMs, including GPT-3, perform badly. If fine-tuned on a training set, their performance improves greatly but this increase does not hold if the distribution of the test set differs from the one of the training set, highlighting that LLMs rely mostly on non-causal clues.

\subsection{Connex Problems}

Although, it is not the main focus of this section, we take a brief look at the problem of retrieving causal structures by merging several sources of data \cite{DBLP:conf/nips/ClaassenH10, DBLP:journals/jmlr/TillmanS11, DBLP:journals/jmlr/TriantafillouT15, DBLP:journals/jmlr/MooijMC20}. This is known as the \textbf{structural causal marginal problem} \cite{DBLP:conf/icml/GreseleKKKSJ22} (i.e. merging several SCMs sharing some unknown nodes in common).
Some works also attempt to generate causal graphs with cycles \cite{DBLP:conf/nips/RothenhauslerHP15, DBLP:conf/uai/ForreM18, bongers2021foundations, DBLP:conf/clear2/AhsanAZ22} or identify the invariant mechanisms in nonstationary data \cite{DBLP:journals/jmlr/Huang0ZRSGS20, DBLP:journals/corr/abs-2204-00607}.

%% file: causal_inference.tex
\section{Causal Inference}
\label{sec:causal_inference}

Causal structure discovery methods can recover the links between variables but does not provide a way to estimate the causal quantities and explain how variables of interest causally affect each others. We describe causal inference as the problem of computing causal effects (for observational, interventional or counterfactual events) or building the functions that can compute them. In this section, we provide extensive overview and explanations of causal inference methods.

\subsection{Do-calculus}
\label{sec:do_calculus}

The \textit{do-calculus} \cite{pearl-2009} establishes inference rules for reducing interventional queries to a probabilistic formulation solvable with observations only.
Do-calculus does not directly provide a computation model for causal inference but is a tool that can be used in conjunction with such a model to estimate causal effects. Do-calculus can be applied once the causal structure of the problem at hand has been found.
The three rules of do-calculus are described in Equations \ref{eq:do_rule1}, \ref{eq:do_rule2}, and \ref{eq:do_rule3}.

\begin{itemize}
    \item \textbf{Rule 1} Deletion of observation\\
    \begin{equation}
        P(y|do(x),z, w) = P(y|do(x), w) \text{~if~} (Y \independent Z|X, W)_{\mathcal{G}_{\overline{X}}}
        \label{eq:do_rule1}
    \end{equation}
    \item \textbf{Rule 2} Reduction of intervention to observation\\
    \begin{equation}
        P(y|do(x),do(z), w) = P(y|do(x), z, w) \text{~if~} (Y \independent Z|X, W)_{\mathcal{G}_{\overline{X}\underline{Z}}}
        \label{eq:do_rule2}
    \end{equation}
    \item \textbf{Rule 3} Deletion of intervention\\
    \begin{equation}
        P(y|do(x),do(z), w) = P(y|do(x), w) \text{~if~} (Y \independent Z|X, W)_{\mathcal{G}_{\overline{X},\overline{Z(W)}}}
        \label{eq:do_rule3}
    \end{equation}
\end{itemize}

\noindent where $\mathcal{G}$ is the causal DAG describing the structure of the problem. The reduction rules are based on the structure of $\mathcal{G}$ and \textbf{d-separation} \cite{DBLP:books/daglib/0066829}. 
A node $Z$ (or a set of nodes, representing causal variables) \textit{d-separates} (or blocks the path between) two nodes $X$ and $Y$ if the latter are independent conditioned on $Z$ (for all distributions represented by the graph $\mathcal{G}$). 

\begin{figure}
    \centering
    \begin{subfigure}{0.5\textwidth}
        \centering
        \includegraphics[scale=0.8]{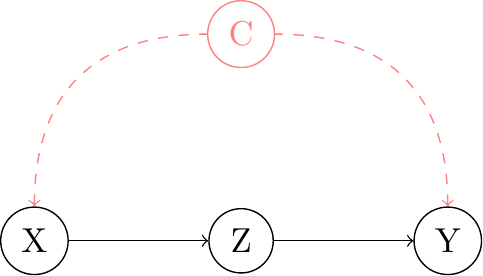}
        \caption{Front-door path}
        \Description[Front door path example]{Example of front door path. A node X is connected to Z, itself connected to Y. A confounder C is connected to both X and Y.}
        \label{fig:frontdoor_path}
    \end{subfigure}
    \hfill
    \begin{subfigure}{0.45\textwidth}
        \centering
        \includegraphics[scale=0.8]{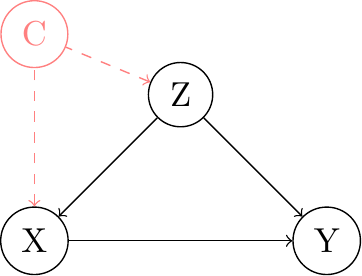}
        \caption{Back-door path}
        \Description[Back door path example]{Example of back door path. A node X is connected to Y and a node Z is connected to X and Y. A confounder C is connected to both X and Z.}
        \label{fig:backdoor_path}
    \end{subfigure}
    \caption{Examples of front-door and back-door paths. The hidden confounder $C$ is not known but the query $P(y|do(x))$ can be answered nonetheless using do-calculus under these graph structures. }
    \label{fig:backfront_paths}
\end{figure}

Rule 1 (\ref{eq:do_rule1}) states that the observation of $Z$ can be ignored if it is independent from $Y$ given some background interventions (on $X$) and observations ($W$). $\mathcal{G}_{\overline{X}}$ is the graph $\mathcal{G}$ under intervention on $X$, i.e. with the incoming links of $X$ removed. We note that $X$ and $W$ can also be (potentially empty) sets of variables.
Rule 2 (\ref{eq:do_rule2}) states that intervening on $Z$ is the same as observing $Z$ if they do not share a common cause (or share a cause already d-separated). If removing directed paths from $Z$ to $Y$ ($\mathcal{G}_{\underline{Z}}$ removes the outgoing links of $Z$) makes them independent, it means that the only path from $Z$ to $Y$ is directed. As the $\mathbf{do}(Z)$ operation forces the value of $Z$ and disconnects it from its causes, the intervention has no impact.
Rule 3 (\ref{eq:do_rule3}) states that an intervention on $Z$ can be ignored if there is not directed path from $Z$ to $Y$. $\mathcal{G}_{\overline{Z(W)}}$ removes the incoming links of $Z$ if $Z$ is \textit{not} an ancestor of $W$. This is to avoid catching $Y \rightarrow Z$ paths.
These rules allow us to estimate the values of some variables under interventional settings using observations only. Whether or not the query can be answered depends on the local shape of the graph. A query is identifiable if and only if it can be reduced to observations using the three rules of do-calculus \cite{DBLP:conf/uai/HuangV06}. Many identifiability and unidentifiability patterns have been discussed in \citet{pearl-2009}. The best known are referred to as the \textbf{front-door} and \textbf{back-door} paths, illustrated in Figure \ref{fig:backfront_paths}.

Under the front-door criterion, the do-calculus rules can be derived into the following formula:

\begin{equation}
    P(y|do(x)) = \sum_{z \in Z} P(z|x) \sum_{x' \in X} P(y|x', z) P(x')
    \label{eq:front_door_formula}
\end{equation}

And under the back-door criterion, the do-calculus rules can be derived into the following formula:

\begin{equation}
    P(y|do(x)) = \sum_{z \in Z} P(y|x, z)P(z) 
    \label{eq:back_door_formula}
\end{equation}

\begin{figure}
    \begin{subfigure}{0.5\textwidth}
        \centering
        \includegraphics[scale=0.8]{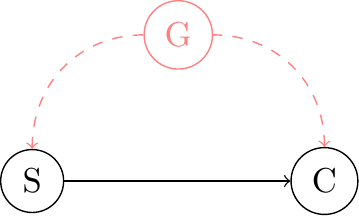}
        \caption{Unidentifiable model}
        \label{fig:smoke_ex_unidentifiable}
        \Description[Unidentifiable causal graph example]{Example of unidentifiable causal graph. Node S is connected to C but a confounder G is connected to both S and C.}
    \end{subfigure}
    \hfill
    \begin{subfigure}{0.45\textwidth}
        \centering
        \includegraphics[scale=0.8]{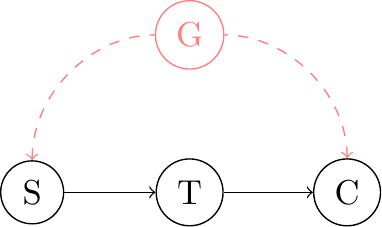}
        \caption{Identifiable (front-door path)}
        \label{fig:smoke_ex_identifiable}
        \Description[Identifiable causal graph example]{Example of identifiable causal graph. Node S is connected T, itself is connected to C. Confounder G is connected to both S and C but not to T.}
    \end{subfigure}
    \caption{Example of causal graph. Smoke (S) is a parent cause of cancer (C), a hidden variable \textcolor{red!50}{(G)}, corresponding to genetic factors, acts as a confounding effect. The model becomes identifiable if a causal link can be made with an intermediate variable (T), tar in the lungs. }
    \label{fig:smoke_ex}
\end{figure}

We continue the example shown on Figure \ref{fig:mr_dag} with now the aim to estimate the values of the variables of interest. Having a potential causal graph linking smoke (S) to cancer (C), can we estimate the effect of smoke on cancer? Or do they have an unobserved genetic common cause \textcolor{red!50}{(G)}?
If no additional information is available (Figure \ref{fig:smoke_ex_unidentifiable}), it is not possible to conclude from observations only. Performing an interventional study (via a Randomized Control Trial) is necessary. However, if we possess additional information, like the presence of tar (T) in the lungs as shown on Figure \ref{fig:smoke_ex_identifiable}, we are in the condition of a front-door path. We can use the formula of Equation \ref{eq:front_door_formula} to solve the query: $P(c|do(s)) = \sum\limits_t P(t|s) \sum\limits_{s'} P(c|s',t) P(s')$, corresponding to the question "what is the probability a patient gets a cancer if he starts smoking?". It can be noted that the ATE or ITE (from the Rubin model) corresponds exactly to the answer we are looking for as it compares $P(c|do(s))$ with $P(c|do(\neg s))$.
The do-calculus can fully answer causal queries, but it requires prior knowledge about the causal graph structure and the data distribution, limiting its application in practice.

\subsection{Re-weighting}
\label{sec:re_weighting}

We now discuss techniques developed to perform causal inference without a graphical model. Instead, we aim to train a model (e.g. with Machine Learning) to compute interventional queries $P(y|do(x))$ from data. 
\textit{Re-weighting} methods assign weights to samples in the data to counter selection bias and get closer to an i.i.d distribution. Cases that happen rarely will be attributed a high weight while cases often occurring will have a low weight.
The implicit assumption behind re-weighting methods is that there is a back-door path (Figure \ref{fig:backdoor_path}) where the $Z$ variable corresponds to the distribution biases. This category of techniques approximates the back-door formula of Equation \ref{eq:back_door_formula} where $P(z)$ is the weight associated with the current sample ($x$, $y$).
Re-weighting approaches can be understood from the perspective of the Rubin model. In this context, the goal is to distribute background variables identically with respect to treatment. For instance, in a medical application, if the data is mostly composed of male patients with few female patients, then the re-weighting will assign a high weight to females. If other background variables are imbalanced, they should be corrected as well. This is typically achieved using a \textit{balancing score} \cite{DBLP:journals/tkdd/YaoCLLGZ21}.
A widely used scoring metric is the \textbf{propensity score} \cite{10.1093/biomet/70.1.41}. It corresponds to the probability of receiving the treatment $X = 1$, given the covariates $z$, and is defined by:

\begin{equation}
    e(z) = P(X = 1|Z = z)
    \label{eq:propensity}
\end{equation}

A common way to assign a weight to each sample is to use an \textbf{Inverse Propensity Weighting} (IPW) estimator \cite{10.1093/biomet/70.1.41}. The weight $w$ of a given sample corresponds to a division by the propensity score if the patient is treated ($X = 1$) and a division by the inverse probability of the propensity score if the patient is not treated ($X = 0$). It is defined by:

\begin{equation}
    w = \dfrac{X}{e(z)} + \dfrac{1 - X}{1 - e(z)}
    \label{eq:ipw}
\end{equation}

Other estimators work on increasing the robustness of the model if the propensity score cannot be easily estimated, like the \textbf{Doubly Robust} (DR) estimator \cite{doi:10.1080/01621459.1994.10476818}, or better estimating the propensity score like the \textbf{Covariate Balancing Propensity Score} (CBPS) \cite{https://doi.org/10.1111/rssb.12027}. However, the problem remains hard to solve in the presence of hidden confounders.

\subsection{Matching}
\label{sec:matching}

\textit{Matching} methods attempt to reproduce the conditions of a randomized trial by dividing the samples into subgroups (or units) of identical covariates. Inference is then performed in a singele subgroup. Each unit in the treated group is matched with another unit in the control group with the same background variables. The intuitive idea behind matching is that, since samples within the same subgroup have the same covariates, only treatment application should differentiate treated and control units \cite{doi:10.1080/00273171.2011.568786, DBLP:journals/tkdd/YaoCLLGZ21}. A popular technique for assigning a sample to a subgroup is the \textbf{Propensity Score Matching} \cite{10.1093/biomet/70.1.41}. The propensity score is used as a distance metric for group assignment: $D(z_i, z_j) = |e(z_i) - e(z_j)|$. This assignment can be performed using nearest-neighbor search, subclassification, or weighting adjustment \cite{stuart2010matching}. However, recent work showed that PSM could create imbalance, model dependence, and bias \cite{king-nielsen-2019}. Other matching methods do not use the propensity score and can alleviate this problem \cite{10.5555/3061053.3061146, iacus-king-porro-2012, DBLP:journals/tkdd/YaoCLLGZ21}.

\subsection{Decision Trees}
\label{sec:trees}

\textit{Decision trees} are another popular structure for causal inference. We follow the Potential Outcome Framework settings and aim to predict the outcome $Y$ of a treatment given background variables $Z=z$. As for matching methods, the principle relies on learning functions operating within a single group of background covariates $z$. A famous example of such models is the \textbf{Classification and Regression Trees} model (CART) \cite{https://doi.org/10.1002/widm.8}, described as $Y = g(z; \mathcal{T}, \mathbf{M}) + \epsilon$, where $\mathbf{M} = \{\theta_1, \dots, \theta_B\}$ is a set of parameters, each associated with one of $B$ terminal nodes of the decision tree $\mathcal{T}$. Regression is performed between $Z$ and $Y$, represented by the linear function $g(\cdot)$. The function is parameterized by the $\theta_i$ values corresponding to the terminal node reached (depending on the $z$ values). $\epsilon \sim N(0, \sigma^2)$ is a noise term. The trees and parameters are learned using \textit{Markov Chain Monte-Carlo} (MCMC) algorithm.
The \textbf{Bayesian Additive Regression Tree} (BART) \cite{10.1214/09-AOAS285} model is an extension of CART based on Random Forests. Instead of relying on a single tree, BART is composed of $T$ trees and sums the output of each one:

\begin{equation}
    Y = \sum\limits_{j=1}^T g(x; \mathcal{T}_j, \mathbf{M}_j) + \epsilon
    \label{eq:bart}
\end{equation}

Relying on the combination of several models reduces the tree sizes. Each tree is much less complex than the original tree in CART methods and behaves as a weak classifier. As several trees are used, BART can incorporate many covariate effects, thus being more robust than single-tree models and yielding great accuracy \cite{DBLP:journals/tkdd/YaoCLLGZ21}.

\subsection{Neural Causal Models}
\label{sec:ncm_inference}

An improvement of the NCM described in Section \ref{sec:neural_causal_model}, called Tractable NCM (TNCM) \cite{DBLP:journals/corr/abs-2110-12052}, replaces the MLPs with \textbf{Sum-Product Networks} (SPNs) \cite{DBLP:conf/uai/PoonD11, DBLP:journals/pami/Sanchez-CaucePD22}. An SPN is a DAG with a single root and with leaf nodes representing a probability distribution over causal variables $\mathbf{X}$. In an SPN, the root node represents the output node. In the context of causal inference, it contains the value $y$ of the query $P(y|x)$. Intermediate nodes represent either \textit{sum} or \textit{product} parameterized operations, alternatively.

The GNN-SCM \cite{DBLP:journals/corr/abs-2109-04173} is another NCM replacing the parametric functions for the causal inference (i.e. the MLPs) with Graph Neural Networks. A Graph Neural Network (GNN) \cite{DBLP:journals/tnn/ScarselliGTHM09, DBLP:conf/iclr/KipfW17} is a particular type of neural network operating on a graph structure. A layer of a GNN takes a graph as input, composed of nodes with feature vectors and an edge list, and returns an aggregation of the node features according to a parametric function and the graph structure. The general form of the GNN equation is described in Equation \ref{eq:gnn_eq}:

\begin{equation}
    \mathbf{h}_V^{(l+1)} = \phi_{\mathbf{\theta_2}} \left(\mathbf{h}_V^{(l)}, \bigoplus_{U \in \mathcal{N}_V} \gamma_{\mathbf{\theta_1}}(\mathbf{h}_V^{(l)}, \mathbf{h}_U^{(l)})  \right)
    \label{eq:gnn_eq}
\end{equation}

\noindent where the vector $\mathbf{h}_V^{(l)}$ represents the node features of node $V$ at layer $(l)$, $\mathcal{N}_V$ is the set of neighbor nodes of $V$ (nodes $U$ with an outgoing edge $U \rightarrow V$), and $\bigoplus$ is the generic aggregation operator. For example,  it can represent a sum $\sum$ or product $\prod$. The GNN layer is composed of the functions $\gamma$ and $\phi$, parametrized by $\theta=\{\theta_1, \theta_2\}$.

The graph structure of the GNN can naturally be used in an SCM: the causal graph corresponds to the input graph of the GNN, and each node is a causal variable. The neighbors of a node $V$ correspond to its parents $\mathbf{pa}_V$ in the DAG and the output of the GNN is $P(V|\mathbf{pa}_V)$.
An intervention on a node $V$ in a GNN can naturally be defined following the GNN formula described above and depriving the set $\mathcal{N}_V$ of the parents of $V$. The new set of neighbors is defined as $\mathcal{M}_V = \{ j | j \in \mathcal{N}_V \setminus \mathbf{pa}_{V} \}$.
As opposed to the above models, the GNN-SCM uses the same GNN mapping function for all node inferences. Only the input graph is modified to match the parents $\mathbf{pa}_{V}$ of the node of interest $V$. However, using a single global function to represent many local mechanisms makes optimization difficult. For this reason, the authors do not consider counterfactual $\mathcal{L}_3$ queries in their work.

\subsection{Variational Graph Auto-Encoders} 
\label{sec:vgae}

The NCM provides a theoretical framework for building neural-based causal models but is not the only way to represent causal relationships with deep neural networks. We present another line of work in this direction, based on Variational Auto-Encoders (VAEs) \cite{DBLP:journals/corr/KingmaW13} and GNNs. VAEs are also widely used for representation learning, as will be explained in Section \ref{sec:representation_learning}.

\begin{figure}
    \centerline{
    \includegraphics[scale=0.58]{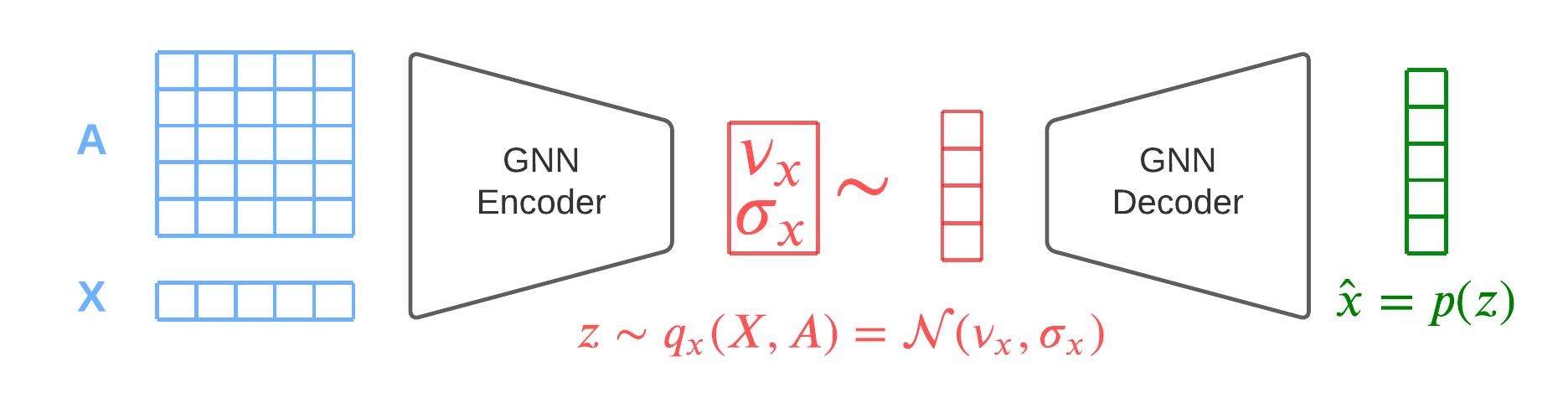}
    }
    \caption{VGAE pipeline. A feature matrix $\mathbf{X}$ and an adjacency matrix $\mathbf{A}$ are fed in input to a GNN encoder to generate the parameters of a multivariate normal distribution, from which the encoded vector $\mathbf{z}$ is sampled. A GNN decoder, fed with $\mathbf{z}$ and $\mathbf{A}$, generates an approximation $\mathbf{\hat{X}}$ reconstructing the original features $\mathbf{X}$. }
    \label{fig:vgae}
    \Description[Illustration of the VGAE process.]{Illustration of the VGAE process. The figure is described in the text. }
\end{figure}

The Variational Autoencoder (VAE) is an encoder-decoder architecture trained by optimizing the \textit{Evidence Lower Bound} (ELBO, a similar loss was briefly described in Section \ref{sec:variational_discovery}). The core idea of a VAE is to encode data into a latent distribution. A decoder samples from this distribution and decodes back the data. The latent distribution is a multivariate normal distribution where the parameters are the output of the encoder. The encoder and decoder are neural networks. Their parameters are learned by optimizing two terms: (i) a data reconstruction term ensures that the output of the decoder is close to the input data and (ii) a regularization term forces the latent distribution to be normal (using Kullback-Leibler divergence).
VAEs are usually applied on grid data like images or tabular data, but \citet{DBLP:journals/corr/KipfW16a} applied the VAE model onto graph structures. This model is called the Variational Graph Auto-Encoder (VGAE) and is described in Figure \ref{fig:vgae}. It is used to generate an embedded dense vector $\mathbf{z}$ of a sparser input feature matrix $\mathbf{X}$. A GNN encoder $q$ learns to generate the distribution of $\mathbf{z} \sim \mathcal{N}(\boldsymbol{\nu}, \boldsymbol{\sigma})$ and a GNN decoder $p$ learns to reconstruct $\mathbf{X}$. 
The encoder is divided into two terms $\boldsymbol{\nu} = \text{GNN}_\nu(\mathbf{X},\mathbf{A})$ and $\boldsymbol{\sigma} = \text{GNN}_\sigma(\mathbf{X},\mathbf{A})$. A VGAE operates similarly as a VAE but with an additional adjacency matrix $\mathbf{A}$ given to the encoders and decoders. Therefore, the encoded $\mathbf{z}$ has the same number $N$ of variables as $\mathbf{X}$ but with fewer dimensions.

\begin{figure}
    \begin{subfigure}{0.45\textwidth}
        \centering
        \includegraphics[scale=0.75]{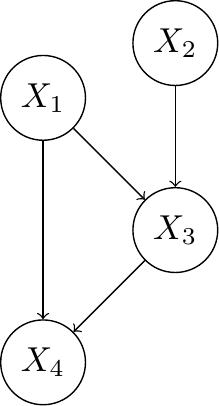}
        \caption{Original graph $\mathcal{G}$. }
        \Description[Illustration of an intervention on a causal graph, original graph]{Illustration of an intervention on a causal graph. The original graph has 4 nodes X1 to X4. X1 is connected to X3 and X4. X2 is connected to X3. X3 is connected to X4. }
    \label{fig:ivgae_inter_g}
    \end{subfigure}
    \hfill
    \begin{subfigure}{0.45\textwidth}
        \centering
        \includegraphics[scale=0.75]{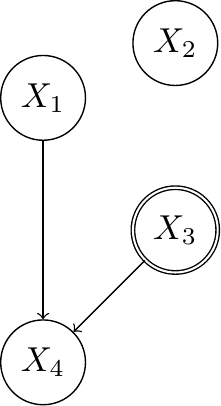}
        \caption{Amputated graph $\mathcal{G}'$. }
        \Description[Illustration of an intervention on a causal graph, amputated graph]{Illustration of an intervention on a causal graph. The amputated graph is similar to the original graph but the links to X3 are removed. }
    \label{fig:ivgae_inter_gprime}
    \end{subfigure}
    \caption{Intervention on the node $X_3$ of graph $\mathcal{G}$. }
    \label{fig:ivgae_inter}
\end{figure}

The interventional VGAE (iVGAE) \cite{DBLP:journals/corr/abs-2109-04173} is a causal model able to perform $\mathcal{L}_2$ causal inference, based on the VGAE architecture. Assuming a causal graph $\mathcal{G}$ (as on Figure \ref{fig:ivgae_inter_g}) and a matrix $\mathbf{X}$ representing the values of the causal variables in $\mathcal{G}$, the reconstruction term $\mathbf{\hat{X}} = p(q(\mathbf{X}, \mathbf{A}_{\mathcal{G}}), \mathbf{A}_{\mathcal{G}})$ computes $\mathcal{L}_1$ queries for every variable $X \in \mathbf{X}$ based on its parents $\mathbf{pa}_X$. This is allowed by the GNN nature of the encoder-decoder, which follows the graph structure when doing inference. Under the assumption that the causal graph is correct, complete, and has no hidden confounders, the iVGAE computes queries $P(y|x)~\forall y \in \mathbf{X}, x \in \mathbf{pa}_Y \subset \mathbf{X}$.
Now, if we amputate the graph $\mathcal{G}$ of precise edges, we are able to generate a \textbf{do} intervention on $\mathcal{G}$. As an example, the graph $\mathcal{G}'$ on Figure \ref{fig:ivgae_inter_gprime}, corresponds do an intervention on $X_3$. Under graph $\mathcal{G}'$, $\mathbf{\hat{X}} = p(q(\mathbf{X},\mathbf{A_{\mathcal{G}'}}), \mathbf{A_{\mathcal{G}'}}) \approx p(\mathbf{X}|\text{do}(X_3=x_3))$.
The iVGAE is a VGAE architecture used as a causal model. Amputating the input graph and forcing the value of the intervened node allows performing $\mathcal{L}_2$ causal inference. 
However, the iVGAE cannot marginalize over exogenous variables and cannot compute $\mathcal{L}_3$ queries.

The Variational Causal Graph Encoder (VACA) \cite{DBLP:conf/aaai/Sanchez-MartinR22} is another model based on a VGAE architecture for causal inference. VACA is defined similarly to the iVGAE but with additional constraints allowing it to perform counterfactual inference, assuming that there are no hidden confounders in the model:
The first constraint is put on the decoder, which \textit{should have at least l-1 hidden layers}, with $l$ being the longest directed path in the causal graph $\mathcal{G}$. It ensures that message propagation in the GNN can transmit the causal effect to every node.
The second constraint is put on the encoder, which \textit{should have no hidden layers}. It ensures that the encoded distribution for each causal variable depends only on its parents.
VACA performs interventions similarly to iVGAE by modifying the causal graph $\mathcal{G}$ into a graph $\mathcal{G}'$. Counterfactual inference is performed by combining two inputs. As shown in Section \ref{sec:pch}, counterfactual requires abduction, action and prediction. A first set of latent variables $\mathbf{Z}_{fact}$ is encoded using the values of $\mathbf{X}$ and the graph $\mathcal{G}$ (abduction), a second set $\mathbf{Z}_{inter}$ is encoded using $\mathbf{X}_{X_i=x_i}$, i.e. $\mathbf{X}$ under the intervention $\text{do}(X_i=x_i)$ on the variable of interest $X_i$ and the modified graph $\mathcal{G}'$. The vector $\mathbf{z}_i$, corresponding to the encoded value of the intervened variable $X_i$, is merged into the set $\mathbf{Z}_{fact}$ to form $\mathbf{Z}$. This new set is then given to the decoder, which returns the expected value for each causal variable (prediction).
If the constraints described above and the unconfoundedness assumption are respected, then VACA is a Neural Causal Model able to represent interventional and counterfactual distributions. In the presence of confounders, some queries can still be approximated if the GNNs are expressive enough \cite{DBLP:conf/aaai/Sanchez-MartinR22}.
The Causal Effect Variational Autoencoder (CEVAE) \cite{DBLP:conf/nips/LouizosSMSZW17} and the Identifiable Treatment Conditional VAE (Intact-VAE) \cite{DBLP:journals/corr/abs-2109-15062} are VAE-based models performing causal-effect inference without using graph structure. These methods aim to identify the individual treatment effect (ITE). The causal graph has a simple and known shape $X \rightarrow Y \leftarrow T$ (outcome $Y$ depends on covariates $X$ and treatment $T$) with possible hidden confounders.
The \textbf{Deep End-to-End Causal Inference} (DECI) \cite{DBLP:journals/corr/abs-2202-02195} model is a pipeline performing both causal structure discovery and causal inference. It prevents misalignment issues between the learned causal structure and the estimation task. The model optimizes an ELBO loss based on a prior over the structure of the causal graph $p(\mathcal{G})$ and a generic neural network for encoding the posterior distribution $q_\theta(\mathbf{A}_\mathcal{G})$. Causal inference is computed using a GNN architecture. DECI assumes that the graph is identifiable and does not contain hidden confounders. DECI combines ideas from the VCN, described in Section \ref{sec:variational_discovery}, and VGAE.

\subsection{Transformers}
\label{sec:tf_inference_time}

The \textbf{Causal Transformer} (CT) \cite{DBLP:conf/icml/MelnychukFF22} is a Transformer-based architecture \cite{DBLP:conf/nips/VaswaniSPUJGKP17} for causal inference in time series. It attempts to recover the counterfactual outcome or individual treatment effect (ITE) $\hat{Y_t}$ based on three separate inputs: the past covariates $\mathbf{Z}_t$, the past outcomes $\mathbf{Y}_{t-1}$, and the past interventions or treatments $\mathbf{X}_{t-1}$ with the one to be performed next $X_t$. Each input is fed to a separate self-attention sub-network. All the outputs are then merged into a joint attention network with cross-attention, fed in return to two classifiers trained in an adversarial fashion: the first is trained to predict the outcome $\hat{Y_t}$ and the second is trained to predict the current intervention $X_t$ ($X_t$ is fed only to the first classifier). This adversarial training aims to learn a representation predictive of the outcome but \textit{not} predictive of the intervention as background variables should not contain any information regarding the intervention. Information about the current treatment from covariates or past treatments necessarily comes from a spurious correlation. Adversarial training is used here to mitigate confounding effects.

\subsection{Vector Autoregressive Models}
\label{sec:vars_time}

Different approaches have been used to take advantage of time information for causal inference, the most popular being the \textbf{Vector Autoregressive models} (VARs). Granger Causality is usually used with a VAR model \cite{DBLP:journals/widm/NogueiraPRPG22, DBLP:journals/kais/MoraffahSKBWTRL21}, where the variables are linked following a linear relation. A widely used model is the AutoRegressive Integrated Moving Average (ARIMA). ARIMA is made of three components: the autoregressive (AR) part predicts the next value of a series $X(t)$ based on its previous values and an error term, the moving average (MA) part performs the prediction based on the past error terms of AR to take into account correlation between the noise terms, and the integration (I) part makes a series stationary if its trend was not stable. The components are represented in Equations \ref{eq:arima_ar}, \ref{eq:arima_ma}, and \ref{eq:arima_i}, respectively. 

\begin{align}
    \label{eq:arima_ar}
    X_{AR}(t) &= c_1 + \epsilon_t + \sum\limits_{i=1}^p \alpha_i X(t-i) \\
    \label{eq:arima_ma}
    X_{MA}(t) &= c_2 + \epsilon_t + \sum\limits_{j=1}^q \beta_j \epsilon_{t-j} \\
    \label{eq:arima_i}
    X_I(t) &= X(t) - X(t - d)
\end{align}

\noindent where $\alpha_i$, $\beta_j$, $\epsilon_t$, $c_1$, and $c_2$ are trainable parameters and $p$, $q$, and $d$ are hyperparameters. The three equations can be combined together to create the ARIMA model: the series $X(t)$ is shifted with the I component and the resulting is given to a merged ARMA equation. This decomposition allows representing many architectures by changing (or setting to zero) the coefficients of the AR,I,and MA components.
Other approaches for modelling time-series for causal inference include Dynamic Bayesian Networks, an extension of Bayesian networks for discrete time, Hidden Markov Models or Gaussian processes, as surveyed in \citet{DBLP:journals/kais/MoraffahSKBWTRL21}.

%% file: representation_learning.tex
\section{Causal Representation Learning}
\label{sec:representation_learning}

Section \ref{sec:neural_causal_model} introduced neural causal models, i.e. causal reasoning engines taking advantage of Deep Learning techniques, and the benefits of such models. However, while DL is applied to low-level data in high-dimension, causal reasoning is performed on high-level variables, usually with semantic meaning and in small dimensions \cite{DBLP:journals/corr/abs-2011-15091}. Going from the former to the latter is a challenging task that uses \textit{Representation Learning}, and generating variables that can be used for causal structure discovery and inference (i.e. causal variables) is called \textbf{Causal Representation Learning} \cite{DBLP:journals/pieee/ScholkopfLBKKGB21}. This section investigates the current representation learning techniques for causal variables.

\subsection{Disentanglement}
\label{sec:disentanglement}

As discussed in the previous sections, causal variables represent abstract concepts sparsely connected to other variables. In particular, the ICM principle states that causal mechanisms are independent, and modifying one should not affect the others \cite{DBLP:journals/pieee/ScholkopfLBKKGB21}. \textit{Disentangled features} respect these properties. They also require fewer samples to achieve similar performance, are less sensitive to noise, and are more robust \cite{DBLP:conf/nips/SteenkisteLSB19}. Disentangling the semantic features in the data could therefore be a way to generate high-level causal variables suitable for causal discovery and inference. This section reviews the fundamental properties of disentangled features, the current disentanglement methods, and how they could help causal representation learning.

Despite being a widely studied topic in Deep Learning research, there is no clear definition of disentanglement. Many definitions have been proposed, but none have been widely accepted. For instance, \citet{DBLP:journals/corr/abs-1812-02230} defines a vector representation as disentangled if it can be decomposed into orthogonal subspaces where a symmetric modification in one subspace will not affect the others.
Even if no rigorous definition is available, most of the works on disentanglement share common principles. The key idea of disentanglement shared in the research community is that a disentangled representation should separate the \textbf{factors of variation} affecting the data \cite{DBLP:conf/icml/KimM18, DBLP:conf/iclr/LocatelloBLRGSB19}. They represent independent semantic interpretable information describing the data \cite{DBLP:conf/icml/MathieuRST19, DBLP:conf/nips/ChenLGD18}.
The notion of \textit{local modification} is essential in the disentanglement literature and can be reconciled with the notion of \textit{intervention} in Causality \cite{DBLP:journals/corr/abs-2204-00607}. Modifying (i.e. intervening on) a disentangled variable should not have any effect on the other variables not that do not causally depend on the former. For instance, changing the lighting of a scene in an image should not change the pose, the orientation, or the labels of the objects. This example is common in the literature as the core domain in which disentanglement is studied and applied is \textit{controlled image generation} \cite{DBLP:conf/iclr/HigginsMPBGBML17, DBLP:conf/nips/ChenLGD18}. As the effect of an intervention on a disentangled latent space can be easily seen on the output image, this domain is well suited for developing disentanglement models.

The majority of articles in the domain rely on Variational Autoencoders and propose techniques for constraining the dimensions of the encoded space to be disentangled. The Deep Convolutional Inverse Graphics Network (DC-IGN) \cite{DBLP:journals/corr/Whitney16} is an early work in this direction. In order to force the disentanglement of the latent space, the training samples are divided into mini-batches of nearly identical images with a single difference between them, corresponding to the modification of a single factor of variation (e.g. orientation, color, lighting). For each mini-sample, the parameters of only one dimension are updated, forcing it to represent the only desired effect. This method requires many samples where only the measured feature changes. This requirement is hardly achievable in practice when dealing with real-world datasets. \citet{DBLP:journals/jmlr/AchilleS18} uses the Information-Bottleneck Lagrangian \cite{DBLP:journals/corr/physics-0004057} to reduce the amount of data information in the weights of neural networks and increase the level of invariance and disentanglement of the latent representations. The IB Lagrangian is defined in Equation \ref{eq:ib_lagrangian}, using the measure of Entropy $H$ and Mutual Shared Information $I$ as defined in a previous section in Equations \ref{eq:conditional_entropy} and \ref{eq:mutual_shared_info}:

\begin{equation}
    \mathcal{L} = H(\mathbf{Y}|\mathbf{Z}) + \alpha I(\mathbf{Z};\mathbf{X})
    \label{eq:ib_lagrangian}
\end{equation}

\noindent where $\mathbf{X}$ and $\mathbf{Y}$ represent the sets of random variables for the input and output images, respectively. $\mathbf{Z}$ is the latent space we are trying to disentangle. We can note that $\mathbf{Z}$ serves the same purpose as hidden confounders as it represents the mechanisms underlying the generation of the data $\mathbf{X}$. The first term controls \textbf{sufficiency}, i.e. if $H$ is minimal, then $\mathbf{Y}$ can be fully predicted from $\mathbf{Z}$, and $\mathbf{Z}$ is sufficient to represent $\mathbf{Y}$. The second term controls \textbf{minimality}, i.e. the smaller $I$, the lower amount of information is shared from $\mathbf{X}$ to $\mathbf{Z}$. The main idea of the paper is that minimal and sufficient representations $\mathbf{Z}$ should also be invariant and disentangled. However, the information measure is not a measure of the quality of the representation, and the representation may not be more informative than the input $\mathbf{X}$.

More recent approaches build upon the $\beta$-VAE model \cite{DBLP:conf/iclr/HigginsMPBGBML17}. This model provides a modification to the loss function that a Variational Autoencoder (VAE) learns to maximize, i.e. the \textit{Evidence Lower Bound} (ELBO) \cite{DBLP:journals/corr/KingmaW13}, given in Equation \ref{eq:beta_elbo}:

\begin{equation}
    \mathcal{L}_\beta = \mathbb{E}_{q_\phi} [\log p_\theta(\mathbf{X}|\mathbf{Z}) ] - \beta \text{ KL}(q_\phi(\mathbf{Z}|\mathbf{X}) || P(\mathbf{Z}))
    \label{eq:beta_elbo}
\end{equation}

\noindent where $p_\theta$ is the probability of observing the input data distribution of $\mathbf{X}$ from the decoder output conditioned by the distribution of $\mathbf{Z}$, $q_\phi$ is the distribution of $\mathbf{Z}$ obtained with the encoder from $\mathbf{X}$, and KL is the Kullback-Leibler divergence. $\phi$ and $\theta$ are the learned parameters of the encoder and decoder, respectively. $\beta > 0$ is a hyperparameter. The $\beta$-VAE uses values $\beta > 1$ whereas the case $\beta=1$ corresponds to the classic ELBO loss. The paper does not contain a full explanation regarding why this method should yield a disentangled representation, but subsequent works \cite{DBLP:conf/nips/ChenLGD18, DBLP:conf/icml/MathieuRST19, DBLP:conf/aistats/Esmaeili0JBSPBD19} explain it by decomposing the Kullback–Leibler divergence term. For instance, the $\beta$-TCVAE \cite{DBLP:conf/nips/ChenLGD18} decomposes the KL term into a sum of three components:

\begin{equation}
    \text{KL}(q_\phi(\mathbf{Z}|\mathbf{X}) || p(\mathbf{Z})) = I(\mathbf{Z}; \mathbf{X}) + \text{KL}(q_\phi(\mathbf{Z}) || \prod\limits_j q_\phi(Z_j)) + \sum\limits_j \text{KL}(q_\phi(Z_j) || P(Z_j)) 
    \label{eq:kl_tc}
\end{equation}

\noindent where the first term is called the \textit{Mutual Information} (I) component. It represents the information shared between $\mathbf{X}$ and $\mathbf{Z}$. The second term is the \textit{Total Correlation} (TC) component. It is a popular measure of the degree of independence of the dimensions of $\mathbf{Z}$. The third term is the \textit{dimension wise} KL divergence. It measures if each individual dimension of $\mathbf{Z}$ is different from its prior. The intuition behind $\beta$-TCVAE is that minimizing the Total Correlation $\text{TC}(z) = \text{KL}(q(z) || \prod_j q(z_j))$ increases the level of disentanglement of $\mathbf{Z}$. The distribution of a disentangled representation $\mathbf{Z}$ should be the same as the product of the distribution of its components $Z_j$. The authors show experimentally that granting a higher weight to TC enhances disentanglement. The FactorVAE \cite{DBLP:conf/icml/KimM18} uses a similar decomposition to increase independence across dimensions of $\mathbf{Z}$. Instead of minimizing the TC term, the Disentangled Inferred prior VAE (DIP-VAE) \cite{DBLP:conf/iclr/0001SB18} introduces a regularizer based on the covariance of the two distributions $\mathbf{X}$ and $\mathbf{Z}$.

Another line of work, based on Generative Adversarial Networks, aims to separate variables linked to a label, sometimes called factors or attributes of \textit{interest}, and the \textit{background} variables or \textit{residual} attributes \cite{DBLP:conf/cvpr/HadadWS18, DBLP:conf/iclr/BauZSZTFT19, DBLP:conf/icml/PengHSS19, DBLP:conf/nips/GabbayCH21}. These works do not attempt to identify each variable individually but try to separate them into two separate latent spaces: invariant variables $\mathbf{Z}_I$ (e.g. the object represented) and domain-specific ones $\mathbf{Z}_D$ (e.g. the environment, style, orientation). 

Recently, a set of experiments and theoretical analysis \cite{DBLP:conf/iclr/LocatelloBLRGSB19} showed that learning a disentangled representation in an unsupervised way was impossible without additional assumptions or inductive biases on the nature of the task. The authors prove the existence of an infinite family of functions yielding an entangled $\mathbf{Z}$ while minimizing TC. 

This finding induced a shift from unsupervised disentanglement to semi-supervised methods. The Adaptive-Group VAE (ada-GVAE) \cite{DBLP:conf/icml/LocatelloPRSBT20} is a weakly-supervised method for disentanglement. The authors show that weakly-supervised disentanglement is possible with few assumptions. If the number of modified factors of variations between two samples if small, then it is possible to retrieve the marginal distributions of each factor and learn disentangled representations. Whether or not unsupervised disentanglement is achievable remains, however, debated \cite{DBLP:conf/iclr/LocatelloTBRSB20, DBLP:conf/nips/HoranRW21}. 
Nonetheless, disentanglement remains actively researched in several domains: vision (for controlled image generation, image-to-image translation or object detection) \cite{DBLP:conf/cvpr/LiZ0CHMHWJ21, DBLP:conf/nips/GabbayCH21, DBLP:conf/iccv/WuLHZ021,DBLP:conf/ijcai/GendronWD23}
, Independent Component Analysis (ICA) \cite{DBLP:conf/aistats/KhemakhemKMH20, DBLP:conf/clear2/LachapelleRSEPL22}, medical imaging \cite{DBLP:journals/corr/abs-2108-12043}, or Graph representations \cite{DBLP:conf/icml/Ma0KW019}. Moreover, the recent success of text-to-image diffusion models \cite{DBLP:journals/corr/abs-2204-06125, DBLP:journals/corr/abs-2205-11487} at generating images could help drive forward the field of disentanglement, mainly based on VAE architectures or adversarial networks.
Recent works on disentanglement take their inspiration from other domains like ICA and Causality. The Identifiable VAE (iVAE, not to be confused with the interventional VGAE introduced earlier) \cite{DBLP:conf/aistats/KhemakhemKMH20} and the regularized VAE \cite{DBLP:conf/clear2/LachapelleRSEPL22} propose methods for retrieving both the disentangled latent variables and the sparse causal graph linking them together. They can be linked back to the CEVAE \cite{DBLP:conf/nips/LouizosSMSZW17} and the Intact-VAE \cite{DBLP:journals/corr/abs-2109-15062}, described in Section \ref{sec:vgae}, which build similar ideas to perform causal inference. The DDG network (Disentanglement-constrained Domain Generalization) \cite{DBLP:journals/corr/abs-2111-13839} is a method for disentangling domain attributes and the ones of interest by imposing constraints in a semantic and a variation space. The authors use it to learn true disentangled causal variables and perform interventions on either the semantic or the variation space, which can be used in return for data augmentation and better generalization.
A recent approach named Independent Mechanism Analysis (IMA) combines ICA with the ICM principle \cite{DBLP:conf/nips/GreseleKSSB21}. IMA-based regularization aims to recover and distinguish the true sources responsible for the generation of a mixture of signals, assuming that each source influences the data independently. The method has been shown to work when the aforementioned assumption is violated and recovers the true causes in the presence of confounders. 

\citet{DBLP:journals/pieee/ScholkopfLBKKGB21} postulates that automatic disentanglement is a necessary step towards learning causal mechanisms from data at scale. This claim is debated \cite{DBLP:journals/corr/abs-2011-15091}. However, even if it reveals founded, it is not sufficient, as once high-level disentangled features have been extracted, causal representation learning still requires to recover the causal links between the features. Finding the appropriate features at the right level of granularity (trade-off between precision and abstraction) remains a challenging problem.

\subsection{Data Augmentation}
\label{sec:data_augmentation}

Generating a good representation of reality relies heavily on the observations available to the model. A fundamental hypothesis in Deep Learning is the assumption that the training data is independent and identically distributed (i.i.d). When true, a model can only learn the desired distinguishing features to improve its performance. However, if this assumption is not respected, the model may learn undesired behaviors caused by confounding effects. 
The techniques discussed so far are model-driven. They rely mainly on improvements of the architecture used, allowing relaxations of the i.i.d assumption. On the other hand, data-driven approaches aim to provide better data to the model. In particular \textit{data augmentation} can ensure that the data remain i.i.d. and has been shown to improve the performance of Deep Learning models \cite{DBLP:journals/corr/abs-2202-08325}. In this section, we briefly introduce methods aiming to provide a greater quantity or quality of data for training the models to approximate a causal behavior.

\begin{figure}
    \centering
    \begin{subfigure}{0.45\textwidth}
        \centering
        \includegraphics[scale=0.7]{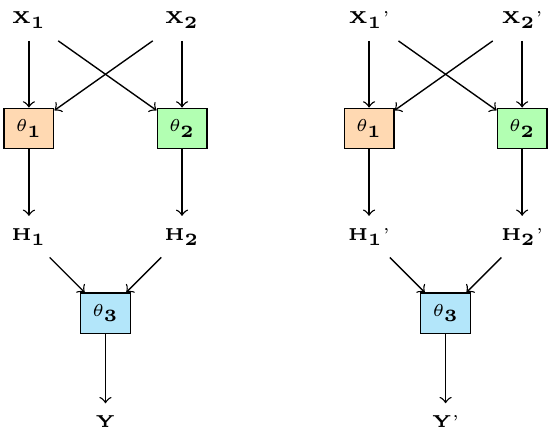}
        \label{fig:itt_inactive}
        \caption{Standard forward pipeline. }
        \Description[Illustration of the forward pipeline in a neural network.]{Illustration of the forward pipeline in a neural network, inputs x1 and x2 are multiplied by both theta1 and theta2. The output of theta1 and theta2 multiplications, called hidden representations H1 and H2, are given to theta3. The output of the theta3 multiplication is the final output Y. The same process is applied to inputs x1' and x2', which generate the hidden representations H1', H2' and the output y'. }
    \end{subfigure}
    \hfill
    \begin{subfigure}{0.45\textwidth}
        \centering
        \includegraphics[scale=0.7]{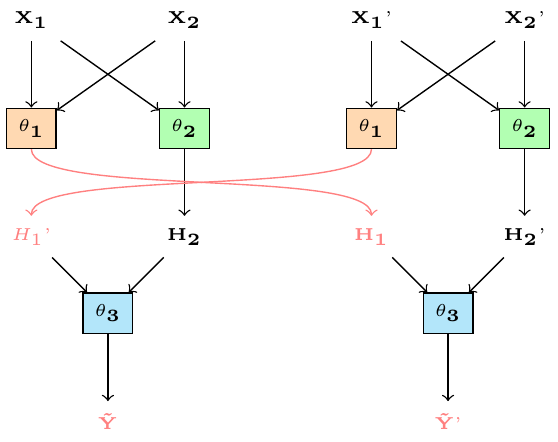}
        \label{fig:itt_active}
        \caption{IIT forward pipeline. }
        \Description[Illustration of the IIT pipeline in a neural network.]{Illustration of the Interchange Intervention Training (IIT) pipeline in a neural network, inputs x1 and x2 are multiplied by both theta1 and theta2 as in the standard pipeline. The hidden representation H1 is exchanged with H1'. Therefore, H1' and H2 are given to theta3. The output of the theta3 multiplication is the final output Y tilde. Similarly, H1 and H2' are given to theta3, which generate the output y' tilde. }
    \end{subfigure}
    \caption{Illustration of Interchange Intervention Training (IIT) in a simple neural network. Two inputs $\mathbf{X} = \{\mathbf{X_1}, \mathbf{X_2}\}$ and $\mathbf{X'} = \{\mathbf{X_1'}, \mathbf{X_2'}\}$ are fed to the same network with parameters $\mathbf{\theta} = \{\mathbf{\theta_1}, \mathbf{\theta_2}, \mathbf{\theta_3}\}$. When performing IIT, a subset of the hidden representations is interchanged before being given to the last layer. Here, $H_1$ and $H_1'$. The outputs after intervention $\mathbf{\tilde{Y}}$ and $\mathbf{\tilde{Y}}$' are compared against the values obtained with a true SCM aligned with the network, fed with the same observational and interventional data. }
    \label{fig:iit}
\end{figure}

The Interchange Intervention Training (IIT) \cite{DBLP:conf/icml/GeigerWLRKIGP22} is a method for training a neural network to match a causal model by feeding it with interventional data. The process consists of two steps: aligning the hidden representations of a neural network with a ground-truth causal model and training the network to match the result of the causal model in a counterfactual setting. IIT mixes two causal situations together to create two counterfactual scenarii running in parallel, as illustrated on Figure \ref{fig:iit}. Two inputs are given to the network, and two hidden representations are exchanged while performing inference, corresponding to performing an intervention in each model. The outputs are compared against the counterfactual outcomes of the ground-truch causal model. The IIT method can be considered a data augmentation method where the augmented data is directly inserted into subparts of the network. However, this training procedure requires knowing the true SCM for the task and decomposing the latent space into several components that can be mapped to causal variables. 
In practice, the true SCM of a given task is unknown, greatly reducing the applicability of this method. 

We make a connection between this work and a data augmentation technique called \textit{ManifoldMixup} \cite{DBLP:conf/iclr/ZhangCDL18, DBLP:conf/icml/VermaLBNMLB19, DBLP:conf/wacv/Mangla0SKBK20}. The latter consist of interpolating the inputs of one neural network layer and the output labels as shown below:

\begin{minipage}{0.45\textwidth}
    \begin{equation}
        \tilde{\mathbf{H_i}} = \alpha \cdot \mathbf{H_i} + (1 - \alpha) \cdot \mathbf{H_i}'
        \label{eq:mixup_input}
    \end{equation}
\end{minipage}
\begin{minipage}{0.45\textwidth}
    \begin{equation}
        \tilde{\mathbf{Y}} = \alpha \cdot \mathbf{Y} + (1 - \alpha) \cdot \mathbf{Y}'
        \label{eq:mixup_label}
    \end{equation}
\end{minipage}

IIT can be seen as a particular case of the ManifoldMixup technique, where $\alpha=0$ in Equation \ref{eq:mixup_input} and the label is determined in a supervised manner using a causal model. The goal of the above techniques is to train a neural model on a more extensive diversity of samples, different than what can be achieved from the training set, to increase model robustness and generalization. It has been shown that ManifoldMixup also increases performance for few-shot tasks \cite{DBLP:conf/wacv/Mangla0SKBK20}. We can argue that the representations generated by the samples correspond to interventions in the neural network. The interchange or mixup of inputs, especially in the hidden layers of the network, yields representations that cannot be reached by simply feeding the network with data \cite{DBLP:conf/icml/VermaLBNMLB19, DBLP:conf/icml/GeigerWLRKIGP22}. This could increase the expressivity of current data augmentation techniques or reduce the amount of samples needed \cite{DBLP:journals/corr/abs-2202-08325}.



%% file: applications_short.tex
\section{Applications}
\label{sec:applications}

The evolution of the Causality field is highly tied to the evolution of science. One of the first occurrences of Causality can be traced back to the second book of \textit{Physics} by Aristotle \cite{falcon2006aristotle}. Causality, therefore, has applications in numerous and broad domains of science, from biology to economics. In this section, we briefly overview the fields using Causality methods. A more extensive review is given in the appendix.

As already discussed in the previous sections, Causality can be applied to improve the performance of Deep Learning models. Some articles have applied Causality theory to improve performance over vision tasks \cite{DBLP:conf/cvpr/QiNHZ20}, complex reasoning tasks \cite{DBLP:journals/kbs/SuiFZCHZ22}, autonomous driving \cite{DBLP:conf/nips/HaanJL19} and causal commonsense \cite{DBLP:conf/semeval/GordonKR12}.
Causality also has applications in biology for modeling dynamic systems such as ecosystems \cite{wright1921correlation, sugihara2012detecting} or attributing gene expression to phenotypical attributes in an individual \cite{yu2004advances, delaplace2010discrete, schadt2005integrative}.
Early Causality theories for time series have been derived for economic applications \cite{granger1969investigating, sims1972money}, particularly econometrics. Those works attempt to use Causality to extract relationships and provide a theoretical economic model. 
Another prominent application of Causality is in finance and the prediction of trends and crises in the stock market \cite{billio2012econometric, papana2017financial}.
As the Granger framework was initially developed for the economy, the Rubin Causal Model was created for medical applications \cite{rubin1974estimating}, particularly for studying the impact of randomized trials. Causal reasoning reduces bias and confounding effects in such studies, and researchers argue that causal methods should be more widely used to reduce spurious correlations that lead to wrong conclusions \cite{hernan2019second}. Causal models also exist in epidemiology to identify the covariates and confounders propagating a disease and to model an epidemic in a population \cite{greenland1999causal, hernan2002causal}.
The primary use of Causality in climatology is for modelling climate events as networks. For instance, \citet{malik2012analysis} uses network representations for studying rainfall in India. Climatology tasks are challenging as many factors are involved, with complex non-linear interactions. Some works pointed out the limitations of correlation-based methods for these tasks \cite{ebert2012causal}.
Causality applications can also be found in ethology \cite{TIMBERLAKE1997107, 7809221, Chen_2020, gendron2023behaviour}, neurosciences \cite{kording2007causal, rohe2015cortical, fornito2013graph}, and social sciences \cite{bramley-time-2018, waismeyer2015causal}.

%% file: conclusion.tex
\section{Conclusion}
\label{sec:conclusion}

Causality has a branching in many fields of science. Recently, a growing interest arose from Deep Learning as it can help tackle critical DL challenges, including strong generalization and complex causal reasoning. This survey provides an in-depth introduction to Causality and the attempts made to combine it with DL to guide researchers that aim to use Causality with Deep Learning.
We describe Causality theories for static and time series data, and survey the various methods used for causal tasks. We provide a deeper look at the most recent approaches that combine Deep Learning with Causality. 
We argue that bridging the gap between the two fields is a way to solve the challenges of generalization, modularity, and interventional and counterfactual reasoning encountered by Deep Learning models. All the methods described in this paper pave the way to building better causal AI models that could solve these challenges, although the perspectives of causality are not limited by the set described here. Causality can for instance provide insight and better understanding of the causal reasoning abilities of existing systems and help create more interpretable models.

%% file: applications.tex
\section{More Applications}
\label{sec:more_applications}

The evolution of the Causality field is highly tied to the evolution of science. One of the first occurrences of Causality can be traced back to the second book of \textit{Physics} by Aristotle \cite{falcon2006aristotle}. Causality, therefore, has applications in numerous and broad domains of science, from biology to economy. In this section, we briefly overview the fields using Causality methods. Table \ref{tab:applications} summarizes the main articles we discuss.

\begin{table}[t]
    \centering
    \caption{Overview of research articles studying applications for Causality. }
    \begin{tabularx}{\linewidth}{>{\hsize=.2\linewidth}X>{\hsize=.5\linewidth}X>{\hsize=.3\linewidth}X}
        \hline
        \textbf{Domain} & \textbf{Research question} & \textbf{Articles}  \\
        \hline
        \multirow{2}{.2\linewidth}{Animal Behavior \ref{sec:app_animal}} & Cause attribution in animal & \cite{TIMBERLAKE1997107, volter2017, doi:10.1152/jn.00046.2020} \\
         & Causal model for animals & \cite{7809221, Chen_2020, gendron2023behaviour} \\
        \hline
        \multirow{5}{.2\linewidth}{Automated Reasoning  \ref{sec:app_reasoning}} & Image recognition & \cite{DBLP:conf/cvpr/QiNHZ20, DBLP:conf/nips/YueZS020, DBLP:conf/cvpr/YueWS0Z21, DBLP:conf/cvpr/YangZQ021} \\
         & Question Answering & \cite{DBLP:conf/wsdm/OhTKIK17, DBLP:journals/kbs/SuiFZCHZ22, DBLP:conf/cvpr/0017JEZZ21} \\
         & Real-world Navigation & \cite{DBLP:conf/nips/HaanJL19, DBLP:conf/nips/VorbachHALR21} \\
         & Causal Commonsense Reasoning & \cite{DBLP:conf/aaaiss/RoemmeleBG11, DBLP:conf/semeval/GordonKR12, DBLP:conf/acl/CaselliV17, DBLP:conf/emnlp/PontiGMLVK20, DBLP:conf/ijcai/LiD0HD20, DBLP:conf/acl/DuDX0022} \\
        \hline
        \multirow{4}{*}{Biology \ref{sec:app_bio}} & Modelling of ecosystems & \cite{wright1921correlation, sugihara2012detecting, laubach2021biologist} \\
         & Causal explanation of genes expression & \cite{yu2004advances, schadt2005integrative, sambo2008cnet, delaplace2010discrete, hlavavckova2015lasso} \\
         & Discovery of common causes linking organism phenotypes & \multirow{2}{*}{\cite{shipley1999testing, austin2007species}} \\
        \hline
        Climatology \ref{sec:app_climate} & Modelling of climate mechanisms & \cite{yamasaki2008climate, malik2012analysis, ebert2012causal, wiedermann2016climate,
        chen2022causation} \\
        \hline
        \multirow{2}{*}{Economics \ref{sec:app_eco}} & Prediction of trends in finance & \cite{billio2012econometric, papana2017financial} \\
         & Modelling of economy &  \cite{haavelmo1943statistical, granger1969investigating, sims1972money, yao2016study} \\
        \hline
        \multirow{2}{*}{Health \ref{sec:app_health}} & Treatment-Effect estimation & \cite{rubin1974estimating, 
        hernan2019second, upadhyaya2021scalable} \\
         & Epidemiology & \cite{greenland1999causal, hernan2002causal} \\ 
        \hline
        \multirow{3}{.2\linewidth}{Neurosciences \ref{sec:app_neuro}} & Emergence of causal inference in the brain & \cite{kording2007causal, 
        shams2010causal,
        rohe2015cortical, FRENCH20208} \\
         & Causal modelling of the brain &  \cite{gu2012causal} \\
         & Connectivity of neural systems &  \cite{seth2005causal, rubinov2010complex, fornito2013graph, seth2015granger} \\
        \hline
        Physics \ref{sec:app_physics} & Modelling of physical mechanisms & \cite{DBLP:journals/corr/abs-1803-07616, goyal-recurrent-2020, DBLP:conf/iclr/YiGLK0TT20, DBLP:conf/iclr/BaradelNMM020, DBLP:conf/ijcai/DuanDFT22} \\
        \hline
        \multirow{2}{.2\linewidth}{Social Sciences \ref{sec:app_social}} & Cause attribution in human & \cite{bramley-time-2018, doi:10.1152/jn.00046.2020, waismeyer2015causal, goddu2020transformations} \\
         & Causal modelling of the brain &  \cite{de2018causal, magnotti2018causal} \\
        \hline
    \end{tabularx}
    \label{tab:applications}
\end{table}

\subsection{Animal Behavior}
\label{sec:app_animal}

A great line of work studies causal reasoning experimentally in animal.
These works do not provide a generic framework for discovering causal relationships but perform a case-per-case analysis or study specific capabilities in humans and animals \cite{TIMBERLAKE1997107, volter2017, doi:10.1152/jn.00046.2020}.
A few articles apply the Causality theory described in Section \ref{sec:causality_theory} to discover the structure of animal behavior. \citet{7809221} uses a method based on Transfer Entropy (optimal Causation Entropy) to model the flow of information within a swarm of midges. 
Similarly, \citet{Chen_2020} uses Causation Entropy to infer the movement of pigeons. \citet{gendron2023behaviour} combines causal discovery for time-series with Graph Neural Networks to build an SCM for meerkat behaviours.

\subsection{Automated Reasoning}
\label{sec:app_reasoning}

As already discussed in the previous sections, Causality can be applied to improve the performance of Deep Learning models. Some articles have applied Causality theory to improve performance over vision tasks: for visual dialog \cite{DBLP:conf/cvpr/QiNHZ20}, for image recognition in few-shot and zero-shot settings \cite{DBLP:conf/nips/YueZS020, DBLP:conf/cvpr/YueWS0Z21}, or for visual question answering \cite{DBLP:conf/cvpr/YangZQ021}.

Many Question Answering tasks in Deep Learning rely on text comprehension, memorization, or common sense. Some tasks additionally require complex reasoning or logical structure. Causality theory can help tackle these tasks. \citet{DBLP:conf/wsdm/OhTKIK17} uses causal attention for why-QA, and \citet{DBLP:journals/kbs/SuiFZCHZ22} uses a causal filter to reduce the number of spurious correlations for Knowledge Graph QA tasks (KGQA). The ACRE dataset \cite{DBLP:conf/cvpr/0017JEZZ21} is a visual reasoning task requiring retrieving causal mechanisms.

Another application for causal reasoning can be found in autonomous driving and navigation. \citet{DBLP:conf/nips/HaanJL19} investigates causal attribution for imitation learning in control games and autonomous cars. \citet{DBLP:conf/nips/VorbachHALR21} attempt to represent the causal structure of an environment with Liquid Time-constant Networks (LTCs) for autonomous navigation with a drone.

Causal Commonsense Reasoning aims to solve a notion of causal reasoning closer to intuitive human understanding. As the information required to solve the queries is not explicitly given, generating a causal graph for this task is hard. SemEval-2010 Task 7 \cite{DBLP:conf/semeval/GordonKR12} is an evaluation dataset for extracting causes and effects in sentences. Despite being published ten years ago, it remains one of the few benchmarks for causal commonsense reasoning. Another dataset we can mention is COPA \cite{DBLP:conf/aaaiss/RoemmeleBG11}, composed of a set of premises and possible causes or effects. A tested system needs to use commonsense to find the right answer. XCOPA is a multilingual extension of COPA \cite{DBLP:conf/emnlp/PontiGMLVK20}. The CausalBank dataset \cite{DBLP:conf/ijcai/LiD0HD20} is a corpus of cause-effect sentences, each based on a causal graph. The Event StoryLine \cite{DBLP:conf/acl/CaselliV17} dataset contains news documents and provides the basis for a task of information extraction and linking. The e-CARE dataset \cite{DBLP:conf/acl/DuDX0022} introduces a set of causal queries: one cause and several potential effect sentences, along with a causal explanation for the true answer. The main difference with COPA is the presence of the causal explanation.

\subsection{Biology}
\label{sec:app_bio}

Biological applications of Causality are multiple. First, Causality can help model dynamic systems such as ecosystems \cite{wright1921correlation, sugihara2012detecting, laubach2021biologist}.
A second highly impactful application is the attribution of a gene expression to a phenotypical attribute in an individual. Some works generate models of regulatory networks for the expression of genes or other biological pathways \cite{yu2004advances, sambo2008cnet, delaplace2010discrete, hlavavckova2015lasso}. Other works attempt to link gene expression with the emergence of diseases like obesity \cite{schadt2005integrative}.
Third, generating causal links between the various traits of a phenotype can inform about the evolutionary process of species and predict species distribution \cite{shipley1999testing, austin2007species}.

\subsection{Climatology}
\label{sec:app_climate}

The primary use of Causality in climatology is for modelling climate events as networks. For instance, \citet{yamasaki2008climate, wiedermann2016climate} 
provide models for studying the impact of El Ni\~{n}o, a periodic flow of warm water in the Pacific Ocean that has important effects on the temperature and the rainfall of the countries in the region. \citet{malik2012analysis} uses network representations for studying rainfall in India, and \citet{ludescher2021network} uses it to forecast extreme weather events.
Climatology tasks are challenging as many factors are involved, with complex non-linear interactions. Some works pointed out the difficulty of accurately attributing causes and the limitations of correlation-based methods \cite{ebert2012causal, chen2022causation}.

\subsection{Economics}
\label{sec:app_eco}

Early Causality theories for time series have been derived for economic applications \cite{haavelmo1943statistical, granger1969investigating, sims1972money}, particularly econometrics. The field of econometrics corresponds to applying statistical methods to economic data to uncover relationships in the data. Those works attempt to use Causality to extract relationships and provide a theoretical economic model. Some works continue to study the theoretical groundings of the economy with causal methods \cite{yao2016study}.
Another prominent application of Causality is in finance and the prediction of trends and crises in the stock market. Popular methods to this end are based on the Granger Causality or measure of Entropy \cite{billio2012econometric, papana2017financial}.

\subsection{Health}
\label{sec:app_health}

As the Granger framework was initially developed for the economy, the Rubin Causal Model was created for medical applications \cite{rubin1974estimating}, particularly for studying the impact of randomized trials. Causal reasoning reduces bias and confounding effects in such studies, and researchers argue that causal methods should be more widely used to reduce spurious correlations that lead to wrong conclusions \cite{hernan2019second}.
The Rubin model is widely used for medical applications but is not the only one.
\citet{upadhyaya2021scalable} proposes an overview of causal structure discovery applications for healthcare.
Causal models also exist in epidemiology to identify the covariates and confounders propagating a disease and to model an epidemic in a population \cite{greenland1999causal, hernan2002causal}.

\subsection{Neurosciences}
\label{sec:app_neuro}

A wide range of studies focus on discovering how the human brain integrates information from sensory inputs and, in particular, how it performs causal inference \cite{kording2007causal,
shams2010causal, rohe2015cortical, FRENCH20208}.
Other works attempt to find causal patterns in the structure of brains and generate a causal model from it \cite{gu2012causal}.
The connectome is a mapping of neural connections in the brain, and connectomics corresponds to the field of research studying the connectome. Understanding the neural mechanisms happening in the connectome can provide a better understanding of the brain. Works attempt to perform a causal connectivity analysis based on Granger Causality  \cite{seth2005causal, rubinov2010complex, fornito2013graph}. In neuroimaging,  studying the changes in the connectome can bring insight into cognition and individual behavior. Some works use causal networks in time series to model these changes \cite{seth2015granger}.

\subsection{Physics}
\label{sec:app_physics}

Causal methods can be used to model the physical mechanisms of the world. Recurrent Independent Mechanisms \cite{goyal-recurrent-2020} use Deep learning to predict the dynamics of physical objects. Benchmarks have been proposed to train systems to model physical reasoning, e.g. prediction of the mechanics of a collision \cite{DBLP:journals/corr/abs-1803-07616, DBLP:conf/iclr/YiGLK0TT20, DBLP:conf/iclr/BaradelNMM020}. They aim to answer interventional or counterfactual queries. Most of these approaches attempt to reproduce intuitive physics in the way humans do \cite{DBLP:conf/ijcai/DuanDFT22}.

\subsection{Social Sciences}
\label{sec:app_social}

As for animals, some works experimentally study the abilities of human beings to perform causal inference. \citet{doi:10.1152/jn.00046.2020} compares humans and monkeys for stimuli localization, \citet{bramley-time-2018} studies how time affects causal attribution, and the last line of work \cite{waismeyer2015causal, goddu2020transformations} looks at how children learn to perform causal reasoning.
Other research areas, on the other hand, attempt to get a deeper understanding of some causal behaviors of the human brain, especially for perception \cite{de2018causal, magnotti2018causal}.